\documentclass[10pt,journal]{IEEEtran}

\usepackage[T1]{fontenc}
\usepackage{amsmath,amssymb}
\usepackage{microtype}
\usepackage{cite}
\usepackage{url}
\usepackage{xcolor}
\usepackage[most]{tcolorbox}
\usepackage{graphicx}
\usepackage{booktabs}
\usepackage{multirow}
\usepackage[caption=false,font=footnotesize]{subfig}
\usepackage[hidelinks]{hyperref}
\usepackage{setspace}
\makeatletter
\renewcommand{\section}{\@startsection{section}{1}{\z@}%
  {2.2ex plus 1ex minus 0.8ex}%
  {0.4ex plus 0.4ex minus 0ex}%
  {\normalfont\normalsize\centering\scshape}}
\renewcommand{\subsection}{\@startsection{subsection}{2}{\z@}%
  {2.2ex plus 1ex minus 0.8ex}%
  {0.4ex plus 0.3ex minus 0ex}%
  {\normalfont\normalsize\itshape}}
\makeatother
\newcommand{\ind}{\mathbf{1}}

\usepackage[symbol]{footmisc}
\renewcommand{\thefootnote}{\fnsymbol{footnote}}
\newcommand \footnoteONLYtext[1]
{
	\let \mybackup \thefootnote
	\let \thefootnote \relax
	\footnotetext{#1}
	\let \thefootnote \mybackup
	\let \mybackup \imareallyundefinedcommand
}

\hypersetup{pdftitle={AIMS: An Agentic AI Framework for Sim-to-Real Multi-Modal ISAC}}

\title{AIMS: An Agentic AI Framework for \\Sim-to-Real Multi-Modal ISAC}
\author{Yijie Bian, Kai Zhang, Wei Guo, Zixin Wang, Shenghui Song, Jun Zhang, Khaled B. Letaief\vspace{-0.3cm}}
 
\begin{document}
\maketitle     

\footnoteONLYtext{
		\hspace*{-1.85em}

		The authors are with the Department of Electronic and Computer Engineering, The Hong Kong University of Science and Technology, Clear Water Bay, Hong Kong (email: ybianaf@connect.ust.hk, eekaizhang@ust.hk, eeweiguo@ust.hk, eewangzx@ust.hk, eeshsong@ust.hk, eejzhang@ust.hk, eekhaled@ust.hk).
    
}

\begin{abstract}
Multi-modal integrated sensing and communication (ISAC) enables environmental
perception and reliable connectivity for intelligent wireless networks.
Data-driven multi-modal ISAC models depend heavily on annotated real-world
data to learn relationships across sensing and wireless observations, thereby
constraining scalable deployment. Although synthetic data generation reduces the burden, adapting existing
simulation pipelines to a target deployment requires consistent scene,
sensing, wireless, and learning configurations, while mismatches among these
coupled components impair sim-to-real transferability. To address the challenge,
we propose an agentic artificial intelligence (AI) framework for sim-to-real
multi-modal ISAC, named AIMS. Given a natural-language deployment request
specifying the target task, deployment conditions, and real-data budget, AIMS
derives a deployment-specific sim-to-real configuration and coordinates its
execution to produce a deployment-specific task model. A two-agent architecture
coordinates scene construction with task learning. A scene construction
agent generates geographically grounded, synchronized sensing and wireless
records from shared physical states, while a scene understanding agent
configures task-relevant modalities and mixture-of-experts (MoE) learning for
zero-shot inference or few-shot adaptation. Structured domain knowledge guides
dependency-aware planning, while validation evidence supports
feedback-driven revision of affected decisions. Experiments on the
real-world DeepSense~6G dataset demonstrate improved vehicle detection and beam
prediction over the considered simulation and fusion baselines. A separate
orchestration benchmark evaluates task interpretation, dependency reasoning,
and feedback-driven replanning across diverse deployment requests, showing
improved plan correctness with structured domain knowledge and validation
feedback.
\end{abstract}


\section{Introduction}    
\label{sec:introduction}  

Sixth-generation (6G) wireless networks are envisioned to {\color{black}integrate}
information delivery, environmental awareness, and native intelligence
~\cite{itu2023imt2030,letaief2019roadmap6g}. Integrated sensing and communication (ISAC) facilitates this vision by synergizing environmental perception and wireless connectivity through shared radio resources and infrastructure~\cite{liu2022isac}.
Beyond resource sharing, ISAC further enables mutual enhancement by using sensing information to adapt communication to the propagation environment, while exploiting communication signals as probing waveforms to infer environmental properties.
As scene geometry and object states shape both sensing observations and radio propagation, heterogeneous sensing data offer complementary insights into the underlying physical environment. Wireless observations capture electromagnetic responses, whereas visual and
positional data provide semantic and spatial context. Data-driven multi-modal
models can learn task-relevant mappings from these complementary observations
to sensing and communication outputs, supporting tasks such as object
detection and beam prediction.

However, training such models for a target deployment requires physically
aligned sensing observations and task supervision. In particular, cross-modal
learning relies on sensing records associated with consistent physical states,
while communication tasks additionally require synchronized wireless
measurements. Acquiring these data requires coordinated sensor deployment,
calibration, and synchronization, together with wireless measurement and task
annotation. Consequently, these requirements increase the site-specific data
collection effort. Datasets such as DeepSense~6G provide a valuable benchmark
for model development~\cite{alkhateeb2023deepsense}, yet extending coverage
to new environments or modified sensing and wireless configurations requires
additional site-specific measurements. These overheads motivate scalable
methods that reduce repeated real-world data collection while retaining
essential deployment-specific information.

Sim-to-real learning combines synthetic data with sparse real-world
measurements for target-domain adaptation
~\cite{jiang2023digitaltwin,chen2025synthsomtwin,Divyadharshini2025smart}.
Although synthetic data reduce the measurement burden, adapting synthetic-data
and digital-twin pipelines to a target deployment still requires coordinated
scene, sensing, wireless, and learning configurations. Scene geometry and
object kinematics jointly govern sensing observations and radio propagation,
while sensor and radio settings determine their learning representations.
Structural mismatches in these coupled relationships cannot be resolved by
increasing the synthetic sample size alone. Therefore, effective sim-to-real
transfer requires physically aligned synthetic observations and supervision,
together with adaptation to the residual domain gap.

\textcolor{black}{Consequently, deployment adaptation introduces a configuration
challenge beyond accurate physical simulation. Translating a high-level
natural-language deployment request into a coherent sim-to-real configuration
requires interpreting its task and deployment conditions for scene
construction, sensing, and wireless data generation. These conditions
determine the observations and supervision available for task learning, while
the specified real-data budget determines the applicable transfer setting.
Since these components are coupled through shared physical states and
dependencies, changes in deployment assumptions can invalidate different
downstream products. The central challenge is therefore to determine the
required capabilities, identify reusable intermediate products, and revise
affected decisions using validation evidence. Agentic artificial intelligence
(AI) supports this process through requirement interpretation,
dependency-aware planning, and feedback-driven configuration revision.}

\subsection{Related Works}
\label{subsec:related_works}

Environmental observations have increasingly served as out-of-band
information for learning-based wireless decisions. Location-based learning
maps user geometry to beamforming strategies
~\cite{lemagoarou2022locationbeamforming}, while camera observations provide
visual cues for beamforming~\cite{ahn2023visionbeamforming}. Environment
semantics support joint beam and blockage prediction
~\cite{yang2023environmentsemantics}. Light detection and ranging (LiDAR)
point clouds and global positioning system (GPS) measurements
add complementary geometric and positional information for vehicular beam
tracking~\cite{bian2024lidargpsbeamtracking}. More general multi-modal methods
fuse heterogeneous observations and develop structural environment representations for beam tracking and beamforming
~\cite{bian2024sensingbeamtracking,shi2026bemamba,bian2026gsbf}, and mixture-of-experts
(MoE) models coordinate modality-specific representations for ISAC tasks
~\cite{zhang2026moeisac}. These studies establish the value of visual,
geometric, and positional information for communication inference.
{\color{black}These models commonly assume that sensing observations and
wireless supervision have already been collected for a particular deployment.}

{\color{black}Synthetic and digital-twin data reduce repeated site-specific
measurement requirements.} Multimodal-Wireless, Multimodal-NF, and LAMBDA provide
synchronized multi-modal sensing and wireless data for configurable learning
tasks~\cite{mao2026multimodalwireless,li2026multimodalnf,zhou2026lambda}, while DeepMIMO provides
large-scale wireless data from predefined ray-tracing scenarios
~\cite{alkhateeb2019deepmimogenericdeeplearning}. {\color{black}These datasets support
reproducible learning and scalable data generation. Their predefined
environments can differ from a target deployment in geometry and sensing
configuration. Mobility and radio settings can also differ.} Digital-twin reconstruction and real-scene alignment narrow this gap. Digital-twin-based beam prediction has demonstrated synthetic
training followed by adaptation with limited real measurements
~\cite{jiang2023digitaltwin}. SMART associates sensing and communication data with a physical scene, with a
relatively limited range of sensing modalities and communication outputs
~\cite{Divyadharshini2025smart}. SynthSoM-Twin further reconstructs static and
dynamic scene components and coordinates multi-modal sensing with propagation
simulation for sim-to-real learning~\cite{chen2025synthsomtwin}. {\color{black}This closer alignment improves the relevance of synthetic data. Adapting
the pipeline to a new site or task still requires expert coordination for
scene reconstruction and calibration. Cross-modal alignment and simulator
integration add further work.}

Large language models (LLMs) and agent-based methods offer mechanisms for intelligently
organizing wireless knowledge and tools for coordinated execution. WirelessLLM
incorporates wireless knowledge and capabilities into LLMs for domain
reasoning~\cite{shao2024wirelessllm}. {\color{black}WirelessAgent combines
perception and memory with planning and action to connect wireless requests
with domain tools and network operations~\cite{tong2026wirelessagent}.
ComAgent uses specialized LLM agents for problem decomposition and planning,
followed by tool-assisted execution and feedback-based
refinement~\cite{li2026comagent},} while
WirelessAgent++ studies automated agentic workflow design and
benchmarking~\cite{tong2026wirelessagentpp}. 
RadioSim Agent integrates LLM-based orchestration with deterministic
electromagnetic solvers for interactive radio-map generation and
physics-grounded propagation analysis, while AutoNetSim further explores
intent-driven construction of executable three-dimensional (3D) radio environments
and wireless simulations through self-evolving agents
~\cite{hussain2026radiosim,si2026autonetsim}.
Recent tool-augmented agents further externalize domain expertise into
verifiable computational tools and use compact language models primarily for
reasoning and orchestration~\cite{zhang2026maintained}. These studies
{\color{black}demonstrate knowledge grounding and planning together with tool
use and execution feedback. Their applications span wireless reasoning,
network operation, and simulation-based optimization.}

\textcolor{black}{For deployment-specific sim-to-real multi-modal ISAC learning, a remaining
technical challenge is to preserve the physical, task, and execution
dependencies that couple scene construction, synchronized sensing and wireless
data generation, and task learning. Changes in deployment conditions further
require selective reuse or regeneration of affected intermediate products
while maintaining cross-domain consistency.}

\subsection{Contributions}
\label{subsec:contributions}

To address these challenges, we propose an agentic AI framework for
sim-to-real multi-modal ISAC (AIMS). AIMS translates a natural-language
deployment request into a deployment-specific sim-to-real configuration and
coordinates the corresponding physical and learning operations to produce a
deployment-specific task model. The request specifies the target task,
deployment conditions, and real-data budget, while AIMS grounds these
requirements in the shared experiment state.
The main contributions are summarized as follows.

\begin{itemize}

\item \textbf{Agentic Sim-to-Real Model Instantiation:}
We address the state-dependent configuration problem in which the required
operations depend jointly on the deployment request and available intermediate
products. The two-agent architecture grounds these conditions in a shared
experiment state and uses structured task, capability, and dependency knowledge to
determine the required capabilities, reusable intermediate products, and execution plan.
Validation evidence updates the state and triggers local revision of affected
decisions as deployment conditions change.

\item \textbf{Domain-Grounded Capabilities and Cross-Domain Alignment:}
We address sensing--wireless inconsistency by developing domain capabilities
for geographic scene grounding, dynamic multi-modal sensing, wireless
projection, and task learning under a shared physical state. Each capability
specifies its required inputs, generated outputs, and validation conditions.
Common physical states, sample references, and configuration provenance are
propagated across these capabilities to preserve correspondence between
multi-modal observations and task supervision.

\item \textbf{Cross-Domain Sim-to-Real Validation:}
We evaluate AIMS on DeepSense~6G using vehicle detection and beam prediction
as representative sensing and communication tasks. Task-level experiments
examine scene, codebook, and fusion mismatches under zero-shot inference and
limited real-domain adaptation. A separate orchestration benchmark over
140 normal and challenging natural-language deployment requests evaluates
task interpretation, dependency handling, artifact reuse, and feedback-driven
replanning, with AIMS achieving higher plan correctness and dependency-handling
performance across both tiers.

\end{itemize}

\subsection{Paper Organization and Notations}
\label{subsec:organization_notation}

The remainder of this paper is organized as follows.
Section~II presents the system model and problem formulation.
Section~III presents the shared agentic orchestration mechanism.
Section~IV presents the scene construction and scene understanding agents.
Section~\ref{sec:simulation_results} reports the experimental setup and
performance evaluation.
Section~VI concludes the paper.

Throughout this paper, boldface lowercase and uppercase letters denote vectors and matrices, respectively, and calligraphic letters denote sets. The sets of real and complex numbers are denoted by $\mathbb{R}$ and $\mathbb{C}$, respectively. The operators $(\cdot)^{\top}$, $(\cdot)^{\mathsf{H}}$, and $\mathbb{E}[\cdot]$ denote transpose, conjugate transpose, and expectation, respectively. Additional notation is defined when first introduced.
\begin{figure*}[t]
  \centering
  \includegraphics[
    pagebox=cropbox,
    clip,
    width=0.99\textwidth,
    keepaspectratio
  ]{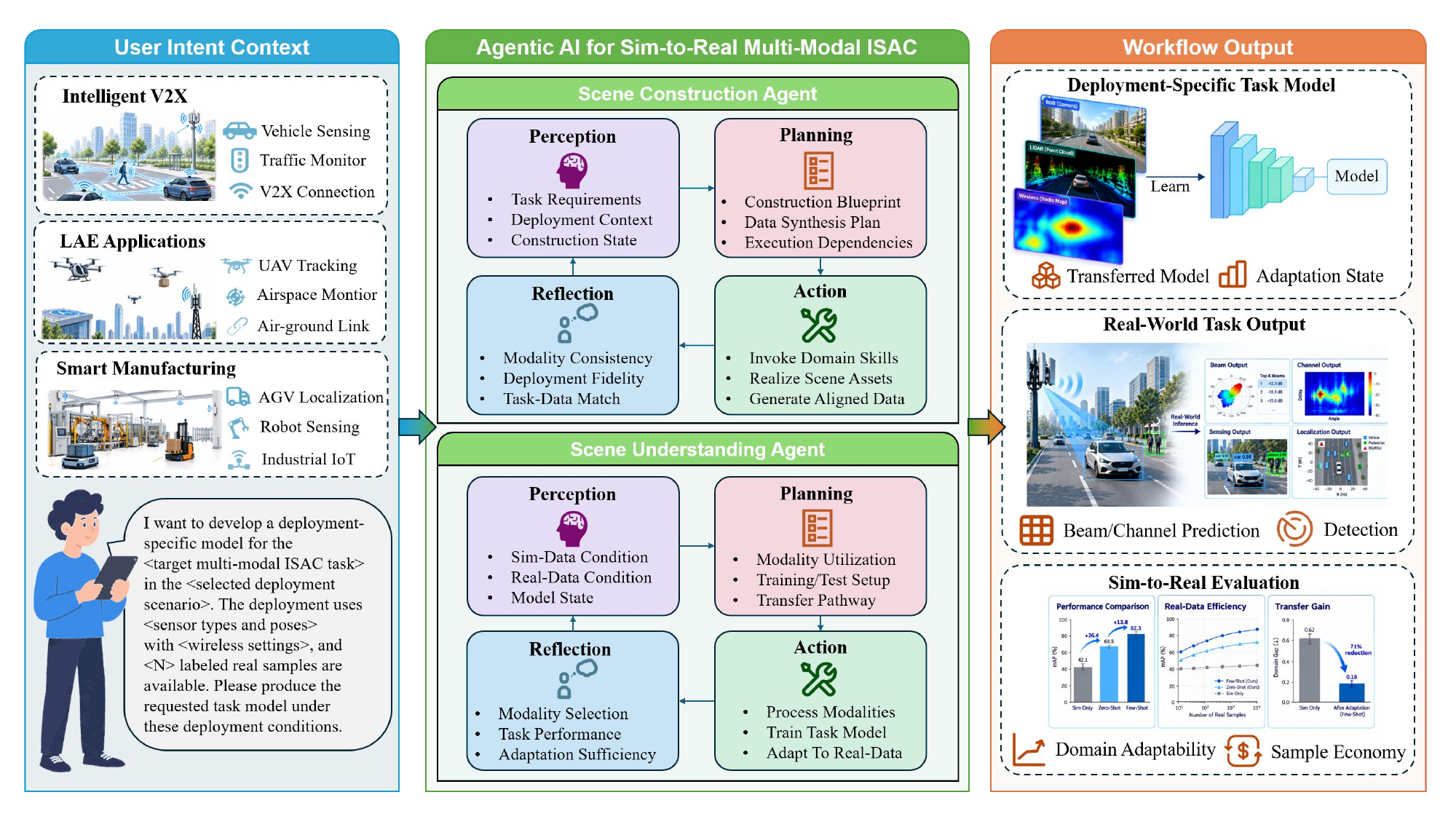}
  \caption{Overview of AIMS for deployment-specific sim-to-real multi-modal
ISAC learning. A natural-language deployment request is organized into a
shared experiment state for two-agent planning and capability execution with
validation-driven revision. The scene construction agent produces aligned sensing
and wireless data, while the scene understanding agent configures learning and
transfer to produce a deployment-specific task model.}
  \label{fig:overall_framework}
\end{figure*}

{\color{black}
\section{System Model and Problem Formulation}
\label{sec:system_model}

This section formulates multi-modal ISAC learning for a target deployment
with limited real supervision. We first establish a common physical
representation for sensing and communication, then formulate the associated
sim-to-real learning problem and deployment-constrained design objective.
The roadside setting provides the representative instantiation evaluated in
Section~\ref{sec:simulation_results}.

\subsection{System Model}
\label{subsec:aims_system_model}

The natural-language deployment request $r$ specifies the target task
$\tau$, user-provided deployment conditions, and the real-data budget
$n_{\text{real}}$. These deployment conditions are organized into a
structured deployment profile for subsequent physical and learning
operations. The corresponding labeled support set
$\mathcal D_{\text{real}}^\tau(n_{\text{real}})$ is available when real-domain
adaptation is requested. The construction specification $\xi$ defines the
synthetic counterpart under the resulting deployment conditions.

Let $\mathcal G$ describe the scene geometry, environmental conditions,
and surface attributes. The surface attributes specify visual appearance
and radio-material properties. At observation instant $t$, the state
$\mathbf s_t$ records object identities, poses, and motion.
The pair $(\mathcal G,\mathbf s_t)$ provides a common physical reference
for sensing observations and wireless propagation. System configurations
determine how these conditions are observed and used for task supervision.

For illustration, we use a roadside vehicular network as a representative
system. The roadside instantiation consists of a roadside unit (RSU) communicating with a target
vehicle while sensors observe the surrounding environment. The RSU provides
downlink transmission, and the communication vehicle serves as the receiving
terminal. This setting instantiates the sensing and communication tasks
evaluated in Section~\ref{sec:simulation_results}.}

\subsubsection{Communication Model}
\label{subsubsec:aims_communication_model}

Consider a transmit node with $N_{\text{t}}$ antennas and a single-antenna
receiving terminal. The roadside instantiation assigns these roles to the
RSU and communication vehicle. The channel of the selected link at
observation instant $t$ is expressed as
\begin{equation}
\mathbf h_t=\sum_{\ell=1}^{L_t}
\beta_{\ell,t}\mathbf a_{\text{t}}(\boldsymbol\vartheta_{\ell,t}),
\label{eq:aims_geometric_channel}
\end{equation}
where $L_t$ is the number of propagation paths, $\beta_{\ell,t}$ is the
complex path coefficient including propagation phase and the fixed
receive response, and $\mathbf a_{\text{t}}(\boldsymbol\vartheta_{\ell,t})$
is the RSU transmit-array response for departure direction
$\boldsymbol\vartheta_{\ell,t}$.
{\color{black}The physical state $(\mathcal G,\mathbf s_t)$ determines the path
coefficients and directions.}
Vehicle motion and surrounding objects therefore alter both the
propagation paths and the preferred transmit beam.

Let $\mathcal F=\{\mathbf f_1,\ldots,\mathbf f_Q\}$ denote the RSU transmit
beam codebook, where $\|\mathbf f_q\|_2=1$.
For a transmitted symbol $u_t$ satisfying $\mathbb E[|u_t|^2]=1$, the
signal received by the vehicle using transmit beam $q$ is
\begin{equation}
r_{t,q}=\sqrt{P_{\text{tx}}} \mathbf h_t^{\mathsf H}\mathbf f_q u_t+n_{t,q},
\label{eq:aims_received_signal}
\end{equation}
where $P_{\text{tx}}$ is the RSU transmit power and
$n_{t,q}\sim\mathcal{CN}(0,\sigma^2)$ is receiver noise.

\subsubsection{Multi-modal Sensing Model}
\label{subsubsec:aims_sensing_model}

Let $\mathcal M_{\text{av}}$ denote the sensing modalities present in the
target deployment and represented by its simulated counterpart.
An observation from modality $m\in\mathcal M_{\text{av}}$ is represented as
\begin{equation}
\mathbf x_t^{(m)}
=
\Psi_m
\left(
\mathcal G,\mathbf s_t;
\boldsymbol\nu_m,\boldsymbol\epsilon_t^{(m)}
\right),
\label{eq:aims_sensor_observation}
\end{equation}
where $\Psi_m$ describes the sensing observation process, $\boldsymbol\nu_m$ contains
the sensor pose and acquisition characteristics, and
$\boldsymbol\epsilon_t^{(m)}$ represents measurement imperfections.
Changes in $\mathcal G$ or $\mathbf s_t$ therefore affect several sensing
observations through their modality-specific measurement processes. Samples
with the same index $t$ correspond to a common dynamic state, while spatial
calibration relates their sensor coordinate frames to the shared scene
reference.

\subsection{Problem Formulation}
\label{subsec:aims_problem_formulation}

The deployment conditions determine the observations and supervision
available for a target task. The formulation connects synthetic-data
construction with task learning under the available real support.

\subsubsection{Task-Specific Multi-modal Inference}
\label{subsubsec:aims_task_formulation}

Let $\tau$ denote a supported sensing or communication task.
Vehicle detection and beam prediction instantiate the task outputs and
learning objectives considered below. 
For task $\tau$, let $\mathcal M_\tau\subseteq\mathcal M_{\text{av}}$
denote the selected sensing modalities and
$\mathcal X_{\tau,t}=\{\mathbf x_t^{(m)}:m\in\mathcal M_\tau\}$ the
corresponding observations. The task model $f_{\boldsymbol{\Theta}_\tau}$ maps
$\mathcal X_{\tau,t}$ and the context vector $\mathbf c_{\tau,t}$ to a
task-specific output. Here, $\boldsymbol{\Theta}_\tau$ denotes the model
parameters, and $\mathbf c_{\tau,t}$ contains task and deployment attributes
available to the model. Detection and beam prediction use separate output
structures and learning objectives.

For vehicle detection, the reference output is the set of vehicles
satisfying the annotation policy in the full image,
\begin{equation}
\mathcal Y_t
=
\left\{
(\mathbf b_{t,j},k_{t,j})
\right\}_{j=1}^{J_t},
\label{eq:aims_detection_target}
\end{equation}
where $\mathbf b_{t,j}\in\mathbb R^4$ denotes the bounding box of the
$j$th vehicle, $k_{t,j}$ denotes its category, and $J_t$ is the number
of annotated vehicles at time $t$.
The detector produces a corresponding prediction
$\widehat{\mathcal Y}_t$ containing object locations, categories, and
confidence scores.
The detection loss
$\ell_{\text{detection}}(\widehat{\mathcal Y}_t,\mathcal Y_t)$ contains the
detector's classification and localization terms.

For beam prediction, the reference beam index is defined by the beamforming
gain under the configured codebook as
\begin{equation}
q_t^\star
=
\underset{q\in\{1,\ldots,Q\}}{\arg\max}
\left|
\mathbf h_t^{\mathsf H}\mathbf f_q
\right|^2.
\label{eq:aims_beam_target}
\end{equation}
The same beam codebook is used to generate synthetic and real beam labels.
The beam-prediction model outputs the probability vector
$\widehat{\boldsymbol\pi}_t=
[\widehat\pi_{t,1},\ldots,\widehat\pi_{t,Q}]^{\top}\in[0,1]^Q$.
The component $\widehat\pi_{t,q}$ represents the predicted probability of
beam $q$, and $\sum_{q=1}^{Q}\widehat\pi_{t,q}=1$.
The predicted beam index is
$\widehat q_t=\arg\max_q\widehat\pi_{t,q}$, and the
classification loss is
$\ell_{\text{beam}}(\widehat{\boldsymbol\pi}_t,q_t^\star)
=-\log\widehat\pi_{t,q_t^\star}$.
{\color{black}The beam index in~\eqref{eq:aims_beam_target} provides
supervision. Inference uses the selected sensing observations.}
The beam-prediction task therefore learns the relationship between
observable scene information and the preferred RSU transmit beam under the
configured codebook.

\subsubsection{Sim-to-Real Training and Adaptation}
\label{subsubsec:aims_sim_real_formulation}

The considered sim-to-real multi-modal ISAC learning problem is specified by
\begin{equation}
    \eta_\tau
    =
    \left\{
    \xi,\,
    \mathcal M_\tau,\,
    \mathcal U_\tau
    \right\},
    \label{eq:aims_experiment_specification}
\end{equation}
where $\xi$ specifies the synthetic construction conditions,
$\mathcal M_\tau$ denotes the selected modality set, and
$\mathcal U_\tau$ contains the parameter groups available for real-domain
adaptation. The specification $\eta_\tau$ is the deployment-specific
sim-to-real configuration instantiated by AIMS from the request $r$, shared experiment state, and structured domain knowledge. The request supplies the target task
$\tau$, deployment conditions, and real-data budget. The experiment state
records the corresponding deployment profile and real-support status.

For each instantiated problem specification $\eta_\tau$, synthetic-domain
training produces
\begin{equation}
    \boldsymbol{\Theta}_\tau^{\text{sim}}
    =
    \operatorname{Train}_{\mathcal M_\tau}
    \left(
    \mathcal D_{\text{sim}}^\tau(\xi)
    \right).
    \label{eq:aims_sim_training}
\end{equation}
The empirical task loss used for numerical learning is
\begin{equation}
\begin{aligned}
    \widehat{\mathcal L}_{\tau}
    (\boldsymbol{\Theta},\mathcal D)
    ={}&
    \frac{1}{|\mathcal D|}
    \sum_{(\mathcal X_\tau,\mathbf c_\tau,Y_\tau)\in\mathcal D}
    \ell_\tau\!\left(
    f_{\boldsymbol{\Theta}}(\mathcal X_\tau,\mathbf c_\tau),
    Y_\tau
    \right).
\end{aligned}
\label{eq:aims_empirical_learning}
\end{equation}
Synthetic pretraining minimizes
\eqref{eq:aims_empirical_learning} over
$\mathcal D_{\text{sim}}^\tau(\xi)$.

In this work, we consider zero-shot inference and real-domain adaptation.
For $n_{\text{real}}=0$, the synthetic-domain model is directly used as $\boldsymbol{\Theta}_\tau^{\text{real}} = \boldsymbol{\Theta}_\tau^{\text{sim}}$.
For $n_{\text{real}}>0$, adaptation initializes from
$\boldsymbol{\Theta}_\tau^{\text{sim}}$ and minimizes
\eqref{eq:aims_empirical_learning} over
$\mathcal D_{\text{real}}^\tau(n_{\text{real}})$. Parameter updates are
restricted to $\mathcal U_\tau$, while the remaining parameters retain their
synthetic-domain values. The optimizer and training settings follow
Section~\ref{subsec:simulation_setup}.

The target-domain risk of the resulting model is
\begin{equation}
    \mathcal R_{\text{real}}^\tau(\eta_\tau)
    =
    \mathbb E_{\text{real}}\!\left[
    \ell_\tau\!\left(
    f_{\boldsymbol{\Theta}_\tau^{\text{real}}}
    (\mathcal X_\tau,\mathbf c_\tau),
    Y_\tau
    \right)
    \right].
\label{eq:aims_target_risk}
\end{equation}
The expectation is taken over task records from the fixed target deployment.
Here, $\mathcal X_\tau$ contains the observations selected by
$\mathcal M_\tau$, $\mathbf c_\tau$ denotes the task context, and $Y_\tau$
denotes the reference output. The target is the annotated vehicle set for
detection or the measured beam index for beam prediction.

\subsubsection{Deployment-Constrained Learning Objective}
\label{subsubsec:aims_deployment_objective}

For task-model deployment, let $\pi$ denote an execution plan and let
$\mathbf z$ summarize the shared experiment state. The state records the
structured deployment profile and real-support status. It also retains
validated products with their configuration references and validation evidence.
The deployment-level design problem is
\begin{equation}
\begin{aligned}
    \underset{\eta_\tau,\pi}{\operatorname{minimize}}
    \quad & \mathcal R_{\text{real}}^\tau(\eta_\tau)\\
    \text{subject to}\quad
    & C_{\text{phy}}(\xi,\pi,r)=1,\\
    & C_{\text{task}}(\eta_\tau,r,n_{\text{real}})=1,\\
    & C_{\text{exec}}(\pi,\eta_\tau,\mathbf z)=1.
\end{aligned}
\label{eq:aims_joint_problem}
\end{equation}
The objective characterizes predictive performance on the target deployment
after synthetic training and any permitted adaptation. The three binary
conditions specify physical consistency, task compatibility, and execution
feasibility.

The physical condition $C_{\text{phy}}$ checks agreement with the declared
scene, sensing setup, and radio configuration. All task-required products
share the physical reference $(\mathcal G,\mathbf s_t)$. For beam prediction,
this reference associates sensing observations with wireless supervision. For
detection, it associates image observations with object annotations. This
condition establishes sample correspondence within the adopted physical
models. The task condition $C_{\text{task}}$ requires a nonempty modality set
$\mathcal M_\tau\subseteq\mathcal M_{\text{av}}$ and supervision
consistent with the requested task. Beam labels follow the configured
codebook in~\eqref{eq:aims_beam_target}. Real-domain parameter updates follow
$\mathcal U_\tau$ and use the corresponding support set.

The execution condition $C_{\text{exec}}$ checks that the supported capability
invocations realize $\eta_\tau$ from $\mathbf z$ with valid input dependencies.
Previously generated products enter the plan when their validated conditions
remain compatible with the requested experiment.
{\color{black}The formulated deployment-constrained learning problem in
\eqref{eq:aims_joint_problem} defines the population-level objective.
AIMS derives a feasible sim-to-real configuration from the deployment conditions and
validation evidence. For the instantiated $\eta_\tau$, numerical learning
minimizes the empirical task loss in
\eqref{eq:aims_empirical_learning}. Held-out target data provide the final
performance evaluation.\par}

{\color{black}
\section{Agentic Orchestration for Sim-to-Real Multi-modal ISAC}
\label{sec:agentic_orchestration_aims}

This section develops the knowledge-guided solution procedure for the
deployment-constrained learning problem in
\eqref{eq:aims_joint_problem}, as illustrated in
Fig.~\ref{fig:overall_framework}. Perception first grounds the
natural-language deployment request $r$ into the shared experiment state
$\mathbf z$ using structured domain knowledge $\mathcal K$ and
state-inspection capabilities. Planning then operates on the grounded state
and $\mathcal K$ to instantiate the experiment specification $\eta_\tau$
and generate the execution plan $\pi$.

Physical capabilities execute $\pi$ and generate the data specified by
$\eta_\tau$, while numerical learners minimize
\eqref{eq:aims_empirical_learning} during synthetic pretraining and restricted real-domain adaptation. Validation evidence identifies configuration conflicts or
inconsistent sample correspondence and updates $\mathbf z$ for local revision
of the affected plan decisions. The primary output is a deployment-specific
task model, while the validated experiment state retains the configuration
provenance associated with its construction.

\subsection{Agentic Formulation and Two-Agent Roles}
\label{subsec:agentic_formulation_and_roles}

AIMS uses structured domain knowledge $\mathcal K$ to connect task
requirements with supported operations. Task knowledge identifies the
observations and supervision associated with each objective. Capability
knowledge specifies the accepted inputs and operating conditions.
Dependency knowledge records the upstream scene states and configurations
that determine each product. The planner uses these relations to propose
an experiment plan, and capability validation checks its inputs against
the declared conditions. Given the grounded experiment state $\mathbf z$ and structured domain
knowledge $\mathcal K$, the LLM planner generates
\begin{equation}
    \pi=\Pi(\mathbf z,\mathcal K),
    \label{eq:agentic_plan}
\end{equation}
where $\pi$ records the experiment specification $\eta_\tau$ and the
capability invocations that realize it. Planning uses the available deployment
evidence to resolve these choices. Numerical learning follows the prescribed
synthetic pretraining and real-domain adaptation procedure.

The two-agent architecture follows the physical and learning components
of $\eta_\tau$. The scene construction agent coordinates the physical
conditions in $\xi$ to generate observations and supervision under a
common scene reference. The scene understanding agent specifies
$\mathcal M_\tau$ and applies the task protocol to configure
$\mathcal U_\tau$. The two agents perform reasoning and coordination over
these decisions, while the corresponding domain capabilities execute the
physical and learning operations specified by the plan.

Before data generation, the scene understanding agent records the required
observations and supervision in $\mathbf z$. The scene construction agent
resolves these requirements using supported capabilities and existing
validated products. The resulting data and configuration references return
through $\mathbf z$ as inputs to learning. Validation evidence identifies
unmet conditions and returns the affected decisions to planning. This
shared-state exchange connects task requirements with their physical
realization throughout the experiment.

\subsection{Perception}
\label{subsec:perception}

Perception grounds the natural-language deployment request $r$ in the shared
experiment state $\mathbf z$. The request specifies the target objective,
deployment conditions, and real-data budget. Perception organizes these
conditions into the structured deployment profile. Before planning,
state-inspection capabilities update $\mathbf z$ with available products,
configuration references, and validation status. The state also records the
real-support status.

Structured domain knowledge $\mathcal K$ resolves the relations represented
in the experiment state. Task knowledge identifies the observations and
supervision required by the target objective. Capability knowledge specifies
supported operations and their input conditions. Dependency knowledge relates
generated products to the physical states and configurations from which they
are derived. These relations distinguish deployment conditions that are
already specified from experiment decisions resolved during planning.

At the task level, perception records the configuration references required
to interpret the available products. For beam prediction, these references
include the sensing configuration and beam codebook associated with the
physical state $(\mathcal G,\mathbf s_t)$. For vehicle detection, they relate
the sensing observations to the applicable annotation policy. The available
real support is associated with the same task state, establishing the
conditions used for subsequent transfer planning.

\subsection{Planning}
\label{subsec:planning}

Planning maps the grounded experiment state to the execution plan $\pi$ by
determining the required endpoint, capability invocations, and reuse of validated products.
Validation evidence supports revision of affected decisions during execution.
A vehicle-detection request requires aligned sensing observations and
annotations, while a beam-prediction request additionally requires wireless
supervision for the same physical samples. Requests for deployment-specific
models extend the plan to learning and transfer operations.

Dependency knowledge determines how deployment changes affect the current
plan. A beam-codebook change modifies the supervision definition while
preserving a channel realization whose scene and radio conditions remain
compatible. A trajectory change modifies $\mathbf s_t$ and invalidates the
dependent sensing and wireless realizations. Planning propagates each change
through the recorded dependencies and retains products whose defining
conditions remain valid.

The scene understanding plan determines the learning configuration from the
validated products. User-specified modalities are retained when supported by
the deployment, while unspecified inputs are resolved from the task
requirements and available sensing configuration. The real support condition
determines whether zero-shot inference or real-domain adaptation is
instantiated. The task protocol specifies the adaptable parameter groups
$\mathcal U_\tau$. These decisions complete the experiment specification
$\eta_\tau$.

\subsection{Action}
\label{subsec:action}

Action realizes the capability invocations selected in $\pi$. The planner
provides each capability with its resolved configuration and required upstream
products. Physical and numerical capabilities operate through defined
interfaces and return their outputs to the shared experiment state. The
resulting records preserve the configuration references required by
subsequent operations.

Capability execution also produces structured validation evidence. The
evidence records whether the declared input conditions are satisfied and
whether the generated product remains consistent with its upstream state.
The evidence updates $\mathbf z$ together with the corresponding output,
providing the experiment state used by downstream capabilities and subsequent
reasoning.

For learning actions, the scene understanding agent invokes the numerical
learner with the selected observations and task supervision. Synthetic
pretraining and real-domain adaptation minimize the empirical task loss in
\eqref{eq:aims_empirical_learning} under their respective data conditions.
Updates during adaptation are restricted to $\mathcal U_\tau$. The
domain-specific realization of the physical and learning capabilities is
detailed in Section~\ref{sec:scene_agents}.

\subsection{Reflection}
\label{subsec:reflection}

Reflection evaluates the returned evidence against the physical, task, and
execution conditions of the formulated deployment-constrained learning
problem in \eqref{eq:aims_joint_problem}. A configuration conflict or invalid
dependency identifies the experiment decision requiring revision. The
corresponding evidence is recorded in $\mathbf z$ and returned to the planner
through \eqref{eq:agentic_plan}.

Plan revision follows the dependency relations encoded in $\mathcal K$. A
change in an upstream physical condition invalidates only the downstream
products that depend on the changed state. Products whose defining conditions
remain compatible are retained in the revised experiment. This dependency
localization restricts reconfiguration to the affected part of $\pi$ while
preserving the validated state of the remaining experiment.

Learning-side reflection evaluates compatibility between the available
products and the current learning configuration. Changes in sensing
availability revise the corresponding input configuration, while changes in
real support update the applicable transfer operation under the task
protocol. When the supported capabilities cannot satisfy a requested
condition, AIMS records the unmet requirement as a capability gap. The
next planning step uses the updated experiment state.}

\section{Scene Construction and Understanding Agents}
\label{sec:scene_agents}

\begin{figure*}[t]
  \centering
  \includegraphics[
    pagebox=cropbox,
    clip,
    width=0.99\textwidth,
    keepaspectratio
  ]{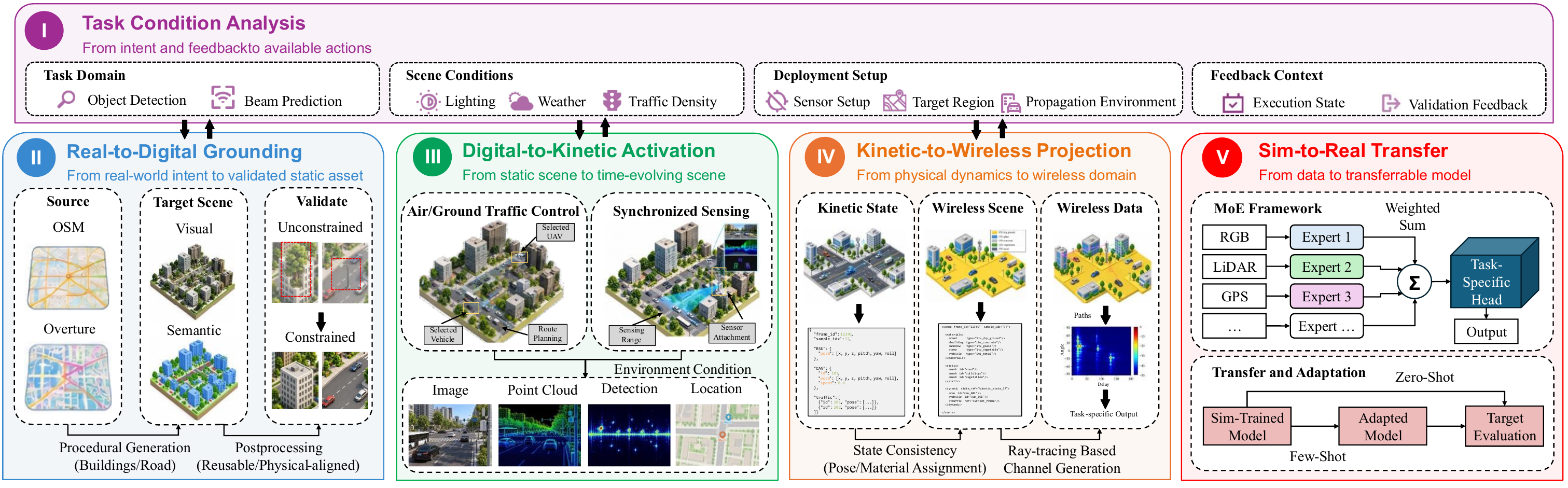}
  \caption{Realization of AIMS. The scene construction agent
coordinates geographic grounding, dynamic sensing, and wireless projection
according to task requirements. The scene understanding agent organizes
multi-modal learning and configures zero-shot inference or few-shot
{\color{black}adaptation according to the available real support.} Execution validation
provides feedback for plan revision. Target evaluation reports predictive
performance on held-out real data.}
  \label{fig:two_agent_implementation}
\end{figure*}

In this section, we instantiate the agentic decisions formulated in
Section~\ref{sec:agentic_orchestration_aims} for sim-to-real multi-modal ISAC
learning, as illustrated in Fig.~\ref{fig:two_agent_implementation}. Each
stage is implemented through a domain capability interface that specifies its
required inputs, generated outputs, and validation conditions, connecting the
agentic decisions in Section~III to the physical and learning operations
described below. We first
organize the grounded task and deployment information into a task condition
shared by the two agents. We then realize task-aligned scene construction and
synchronized sensing and wireless data generation. The resulting records
support multi-modal learning and real-domain transfer.

\subsection{Task and Deployment Instantiation}
\label{subsec:task_deployment_instantiation}

The experiment specification $\eta_\tau$ determined in
Section~\ref{sec:agentic_orchestration_aims} provides the construction and
learning configuration realized by the following capabilities. For vehicle detection, the task condition specifies the
sensing observations and annotation products required by the detector. Beam
prediction additionally specifies wireless supervision for the same physical
samples. {\color{black}The deployment profile supplies the physical scene and
mobility profile. It also specifies the sensing and wireless configurations
used for synthetic-data generation.}

The sensing condition identifies the available modalities and their
acquisition profiles. {\color{black}Representative observations include
red-green-blue (RGB) images and LiDAR point clouds. Radar measurements and GPS
positions provide additional inputs.} The acquisition
profiles define the sensor geometry and sampling behavior. For beam
prediction, the wireless condition also specifies the carrier frequency,
antenna configuration, and beam codebook. A vehicle-detection request may
instantiate RGB or RGB--LiDAR sensing with object annotations. A
beam-prediction request may combine RGB, LiDAR, and GPS observations
with channel or beam-index supervision. User-requested modalities are
retained, and other sensing inputs are resolved from the target deployment.

The instantiated task condition also defines the data and learning endpoints.
It records the requested number and coverage of synthetic samples, the
available real support, and the task outputs required for
learning and transfer. {\color{black}For example, a beam-prediction task may
require synchronized multi-modal observations and channel-derived beam labels.
It can continue through synthetic-domain training and zero-shot inference or
few-shot adaptation at the target deployment. A vehicle-detection task can terminate at
synchronized sensing, annotation generation, and task learning.} The task
condition provides the configuration shared by the scene construction agent
and the scene understanding agent.

\subsection{Scene Construction Agent}
\label{subsec:scene_construction_agent}

Given the instantiated task and deployment conditions, the scene construction
agent realizes the task-aligned synthetic domain through three construction
stages and an alignment handoff. The stages comprise real-to-digital
grounding, digital-to-kinetic activation, and kinetic-to-wireless projection.
Real-to-digital grounding establishes the geographically referenced static
environment and infrastructure. Digital-to-kinetic activation introduces
dynamic objects, trajectories, and synchronized multi-modal sensing
observations. Kinetic-to-wireless projection maps the same dynamic states to
their corresponding radio-domain quantities, including propagation, channel,
and beam-domain products when required by the target task. {\color{black}Finally,
cross-domain alignment links each sensing or wireless product to its annotation
or task supervision through common physical states and sample references. This
alignment produces task-ready synthetic records}
for subsequent learning and sim-to-real transfer.

\subsubsection{Real-to-Digital Grounding}
\label{subsubsec:real_to_digital}

Given the geographic extent and infrastructure conditions specified in the
deployment profile, real-to-digital grounding establishes the
validated static physical reference for subsequent mobility, sensing, and
wireless simulation. The scene construction agent determines
whether the shared experiment state already contains a validated static scene
whose geographic coverage, infrastructure configuration, and semantic
content are consistent with the deployment profile.
A compatible scene is reused as the common spatial reference for the
current task. A new grounding operation is initiated when the requested
region or static-scene conditions are not covered by the existing
validated result.

OpenStreetMap~\cite{mordechai2008osm} provides the primary geographic description, while
Overture Maps supplies complementary attributes and cross-source
evidence for the target area. The selected geographic information is
normalized to a common coordinate reference and transformed into a
three-dimensional representation of the transport network, built environment,
and relevant natural features. The
same coordinate reference is used to interpret the RSU and roadside
sensor poses specified in the deployment profile. Semantic and infrastructure
attributes are retained to support downstream mobility, sensing, and radio
simulation.

Building appearance is instantiated through procedural facade generation
and texture assignment on the reconstructed geometry. The visual attributes
define the surface detail observed under the configured camera and
illumination conditions. Scene semantics associate the rendered objects
with their geometric representations. The radio scene uses the exported
surfaces and their assigned electromagnetic material properties under the
same spatial reference. 

Geometric and semantic validation checks object placement against the
reconstructed building volumes and road geometry. Vegetation intersecting
a building volume is removed, and roadside objects conflicting with the
road layout are corrected or rejected according to the available source
evidence. The checks preserve the spatial relationships used by sensor
rendering and radio-scene construction. Unresolved source ambiguities are
recorded as validation gaps.

The resulting static scene provides the validated spatial reference for
dynamic generation and subsequent wireless projection.

\subsubsection{Digital-to-Kinetic Activation}
\label{subsubsec:digital_to_kinetic}

Digital-to-kinetic activation transforms the validated static scene into
time-varying physical states that support synchronized sensing and subsequent
wireless projection. The scene construction agent first determines whether
the shared experiment state contains a validated kinetic realization whose mobility,
environmental, and sensing conditions remain compatible with the current
deployment request. Compatible trajectories and sensing records can be
reused. {\color{black}Changes in mobility, environment, or sensing configuration
require regeneration of the affected dynamic products.}

Once the required kinetic configuration is established,
CARLA~\cite{dosovitskiy2017carla} instantiates the communication vehicle and
surrounding traffic on the validated road network and evolves the scene
according to the mobility profile and environmental conditions in the
deployment profile. The communication vehicle is tracked separately from
background traffic so that its position, orientation, and motion remain
associated with the corresponding trajectory throughout the experiment. At
each retained observation instant, these actor states form the dynamic state
$\mathbf{s}_t$ introduced in Section~II, which provides the common physical
reference for sensing generation and the subsequent wireless projection.

The sensing configuration follows the sensor geometry and acquisition settings
of the target deployment. {\color{black}From each retained dynamic state, the
enabled RGB and LiDAR modalities generate corresponding observations. Enabled
radar and positioning modalities use the same physical scene state.} Simulated actor identities and geometry also
provide task supervision such as object locations and categories. Data
collection is restricted to the effective sensing region defined by the
deployment configuration, so that the retained records correspond to the
portion of each trajectory that is relevant to the roadside sensing setup.
For vehicle detection, these records associate sensing observations with
object annotations. For beam prediction, the same retained samples also
provide the communication-vehicle state required by the subsequent wireless
projection, thereby preserving the physical correspondence between sensing
inputs and beam supervision.

Before the kinetic state is exposed to downstream operations, {\color{black}trajectory and
sensing validation checks the realized motion and sensor configuration. It
also verifies sample completeness and temporal consistency.} Common actor
and sample references associate observations with annotations generated from
the same dynamic state. The output is a validated kinetic representation that
combines synchronized physical states with sensing observations and task
annotations. This representation provides the dynamic physical
interface used by both the scene understanding agent and the
kinetic-to-wireless projection stage.

\subsubsection{Kinetic-to-Wireless Projection}
\label{subsubsec:kinetic_to_wireless}

Kinetic-to-wireless projection maps each validated kinetic sample to the
radio-domain representation required by the target task. The scene
construction agent invokes this stage only when wireless supervision or
intermediate radio information is required. It first determines whether a
compatible wireless result already exists for the same physical state and
radio configuration. The validated static scene and the dynamic state
$\mathbf{s}_t$ define the relative geometry of the communication nodes and
surrounding objects, while the deployment profile specifies the carrier and
antenna settings. The resulting wireless realization therefore remains tied
to the same physical sample used for sensing generation.

In the current realization, Blender converts the grounded kinetic scene from CARLA into
a ray-tracing representation while preserving the common coordinate reference
established in the previous stages. Dynamic objects and communication nodes
are instantiated from the retained kinetic state, and supported surfaces are
assigned radio-material properties. The Sionna RT ray-tracing
engine~\cite{hoydis2023sionna} then
evaluates the propagation paths associated with each retained sample and
derives the corresponding channel response from the scene geometry, material
properties, and radio configuration. The resulting path information comprises
complex path parameters and the assembled channel response. Line-of-sight and
non-line-of-sight conditions arise directly
from the instantiated scene geometry.

The endpoint of wireless processing is determined by the requested task. A
channel-oriented task exposes the channel representation directly. For beam
prediction, the configured codebook maps each channel response to beam-domain
powers and the corresponding beam label.

Before the wireless result is exposed to the scene understanding agent,
validation checks its consistency with the corresponding kinetic record. The
checks cover geometric, radio, and sample-level consistency. Validated
wireless outputs inherit the
same trajectory and sample references as the sensing records, which preserves
the physical correspondence between sensing observations and wireless
supervision. The output of this stage is therefore a validated wireless
representation that can be combined with the synchronized sensing records for
task learning and sim-to-real transfer.

\subsection{Scene Understanding Agent}
\label{subsec:scene_understanding_agent}

The scene understanding agent converts the validated multi-modal records
produced by the scene construction agent into a deployment-specific task
model. {\color{black}Scene construction determines the physical content of
the synthetic domain. Scene understanding configures the resulting
observations and supervision for task learning and transfer.} The
validated records preserve a common physical reference across sensing and
wireless quantities and constitute
$\mathcal{D}_{\text{sim}}^{\tau}(\xi)$ defined in
Section~\ref{subsubsec:aims_sim_real_formulation}.

For a target task $\tau$, the agent organizes the admissible sensing inputs,
the corresponding learning configuration, and the transfer strategy according
to the deployment context and the available real support. It first
performs task-adaptive multi-modal learning in the synthetic domain and then
determines whether the resulting model is used directly or adapted to the
target deployment. The following subsections describe these two operations.

\subsubsection{Task-Adaptive Multi-modal Learning}
\label{subsubsec:task_adaptive_multi_modal_learning}

The scene understanding agent converts the validated synthetic records into
a task-specific learning configuration. Given a target task $\tau$ and the
sensing modalities $\mathcal{M}_{\text{av}}$ available in the deployment,
the agent specifies the input modality set
$\mathcal{M}_{\tau}\subseteq\mathcal{M}_{\text{av}}$ according to the task
requirements and deployment context. The MoE model learns the contribution
of each selected modality from synthetic data.

For each modality $m \in \mathcal{M}_{\tau}$, let
$\mathbf{x}_{t}^{(m)}$ denote its observation at sample $t$. A
modality-specific encoder $e_m(\cdot)$ extracts its representation, which is
mapped into a task-specific fusion space by $\phi_{m,\tau}(\cdot)$ as
\begin{equation}
    \mathbf{Z}_{m,\tau,t}
    =
    \phi_{m,\tau}
    \left(
        e_m\left(\mathbf{x}_{t}^{(m)}\right)
    \right).
\end{equation}
Let
$\mathcal{Z}_{\tau,t}
=
\{\mathbf{Z}_{m,\tau,t}:m\in\mathcal{M}_{\tau}\}$.
The MoE gating network assigns a normalized weight to each selected modality
according to its current representation and the task context
$\mathbf{c}_{\tau,t}$,
\begin{equation}
    \gamma_{m,\tau,t}
    =
    \frac{
        \exp\left(
            g_{m,\tau}
            \left(
                \mathcal{Z}_{\tau,t},
                \mathbf{c}_{\tau,t}
            \right)
        \right)
    }{
        \sum_{j\in\mathcal{M}_{\tau}}
        \exp\left(
            g_{j,\tau}
            \left(
                \mathcal{Z}_{\tau,t},
                \mathbf{c}_{\tau,t}
            \right)
        \right)
    }.
\end{equation}
The fused task representation is then
\begin{equation}
    \mathbf{F}_{\tau,t}
    =
    \sum_{m\in\mathcal{M}_{\tau}}
    \gamma_{m,\tau,t}\mathbf{Z}_{m,\tau,t}.
\end{equation}

The representation $\mathbf{F}_{\tau,t}$ is provided to the corresponding
task head. For vehicle detection, the modality mappings preserve the spatial
structure required for object localization. For beam prediction, the fused
features form a link-level representation used to predict the beam
distribution. {\color{black}The encoders, gating network, and task head
minimize~\eqref{eq:aims_empirical_learning} over
$\mathcal D_{\text{sim}}^\tau(\xi)$. This procedure produces
$\boldsymbol{\Theta}_\tau^{\text{sim}}$.\par}

The selected modality set and the resulting synthetic-domain model become
part of the learning state maintained for the target deployment. A change in
the target task or in the sensing modalities available for inference requires
the affected learning configuration to be reconsidered.

\subsubsection{Deployment-Specific Sim-to-Real Transfer}
\label{subsubsec:sim_to_real_transfer}

The scene understanding agent instantiates the transfer configuration defined
in Section~\ref{subsubsec:aims_sim_real_formulation}. Given the
synthetic-domain model $\boldsymbol{\Theta}_\tau^{\text{sim}}$, the available
real support determines the applicable transfer branch. For
$n_{\text{real}}=0$, the synthetic-domain model is used directly for
zero-shot inference. For $n_{\text{real}}>0$, the agent invokes real-domain
adaptation using $\mathcal D_{\text{real}}^\tau(n_{\text{real}})$ under the
adaptation scope $\mathcal U_\tau$.

The task protocol determines the parameter groups exposed to adaptation.
Depending on the task implementation, $\mathcal U_\tau$ can include
modality-specific representations, the fusion module, and the task head.
Once the transfer configuration is instantiated, the numerical learner
executes the empirical optimization defined in
\eqref{eq:aims_empirical_learning}. The resulting model
$\boldsymbol{\Theta}_\tau^{\text{real}}$ and its transfer configuration are
returned to the shared experiment state.

When the target task and selected modality set remain unchanged, a change in
the real-data budget leaves the validated synthetic domain applicable to the
experiment. The scene understanding agent therefore updates the transfer
operation while retaining the compatible synthetic records and
$\boldsymbol{\Theta}_\tau^{\text{sim}}$. This dependency preserves the
separation between physical data construction and deployment-specific model
adaptation.

\begin{figure*}[!t]
    \centering

    \subfloat[Synthesized Scenario 3.]{
        \includegraphics[width=0.315\textwidth]{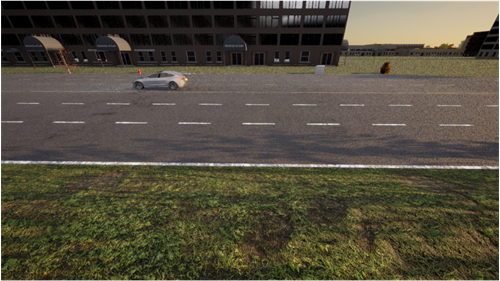}
        \label{fig:scenario3_synthetic}
    }
    \hfill
    \subfloat[Synthesized Scenario 4.]{
        \includegraphics[width=0.315\textwidth]{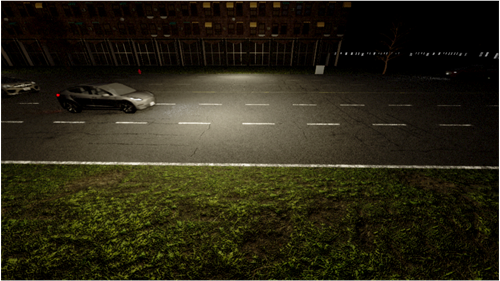}
        \label{fig:scenario4_synthetic}
    }
    \hfill
    \subfloat[Synthesized Scenario 9.]{
        \includegraphics[width=0.315\textwidth]{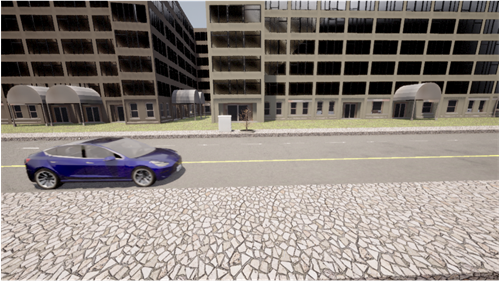}
        \label{fig:scenario9_synthetic}
    }

    \par
    \subfloat[DeepSense 6G Scenario 3.]{
        \includegraphics[width=0.315\textwidth]{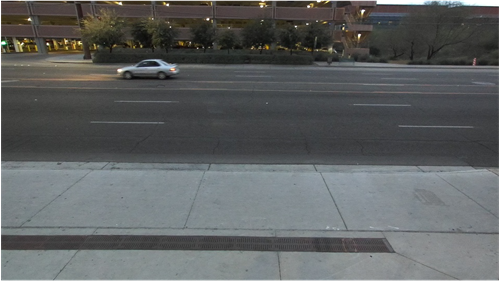}
        \label{fig:scenario3_real}
    }
    \hfill
    \subfloat[DeepSense 6G Scenario 4.]{
        \includegraphics[width=0.315\textwidth]{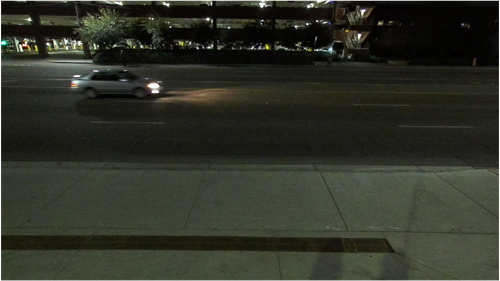}
        \label{fig:scenario4_real}
    }
    \hfill
    \subfloat[DeepSense 6G Scenario 9.]{
        \includegraphics[width=0.315\textwidth]{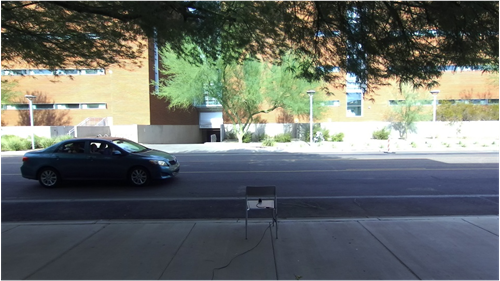}
        \label{fig:scenario9_real}
    }

    \caption{Examples of the reconstructed synthetic environments and the
corresponding real DeepSense 6G deployments for Scenarios 3, 4, and 9.}
    \label{fig:scene_reconstruction_examples}
\end{figure*}

\section{Experimental Results}
\label{sec:simulation_results}

In this section, we evaluate the proposed AIMS framework using the real-world DeepSense~6G multi-modal ISAC dataset~\cite{alkhateeb2023deepsense}.
We consider both sensing and communication tasks, including vehicle detection and beam prediction. We also evaluate the agentic orchestration performance.

\subsection{Experimental Setups}
\label{subsec:simulation_setup}

{\color{black}We reconstruct DeepSense~6G deployment environments for
Scenarios 3, 4, and 9~\cite{alkhateeb2023deepsense}. The reconstructions
generate synthetic sensing and wireless data for sim-to-real learning.}
Scenarios 3 and 4 are used for beam prediction,
while Scenarios 3 and 9 are used for communication-vehicle detection.

\subsubsection{Sensing Settings}
RGB and GPS are used for the quantitative tasks in Scenarios 3, 4, and 9.
The roadside camera provides RGB images with a native resolution of $960\times540$, a field of view of $110^{\circ}$, and a nominal acquisition rate of 30~frames/s.
Vehicle-mounted GPS receivers provide position measurements at 10~Hz.
Scenario~9 additionally provides two-dimensional LiDAR scans with a $360^{\circ}$ field of view and a nominal scanning rate of 10~Hz.
The synthetic environments are reconstructed from geographic information using OpenStreetMap and Overture Maps, with CARLA generating vehicle motion and sensor observations under the corresponding deployment and illumination conditions.

\subsubsection{Wireless Settings}
The considered downlink operates at $60$~GHz. The RSU employs a $16$-element
uniform linear transmit array and a transmit beam codebook of $Q=64$ beams,
while the communication vehicle is modeled as a single-antenna
quasi-omnidirectional receiver. Each real wireless record contains the
measured per-beam received powers, and the beam-prediction label is the beam
index with the highest received power. These labels follow the same beam
identities and ordering used in the downlink model.

For synthetic data generation, the reconstructed geometry and dynamic object states are transferred to Sionna RT through the Blender-based scene representation.
Ray tracing produces propagation paths and channel realizations from the RSU to the communication vehicle.
Beam-domain powers are then computed using the same RSU transmit codebook employed by the target deployment.
Synthetic and real beam labels therefore follow the same beam identities and ordering.

\subsubsection{Learning Settings}
We employ task-specific MoE models. RGB uses a ResNet-18 encoder, planar
LiDAR uses PointNet, and GPS uses a position encoder. Context-conditioned
softmax gating fuses the expert features for 64-class beam prediction and
vehicle detection, with spatial features retained for detection. The models
are trained on synthetic data using Adam with a learning rate of $10^{-3}$ and
a batch size of 64. Beam prediction uses cross-entropy loss, while detection
uses classification and localization losses. {\color{black}Zero-shot evaluation
directly applies the learned models. Few-shot adaptation updates the parameter
groups in $\mathcal U_\tau$ using limited labeled real samples.}

We compare the proposed framework with the following baselines and a real-data training benchmark.

\begin{figure*}[t]
\centering
\subfloat[Top-3 accuracy.]{%
  \includegraphics[width=0.49\textwidth]{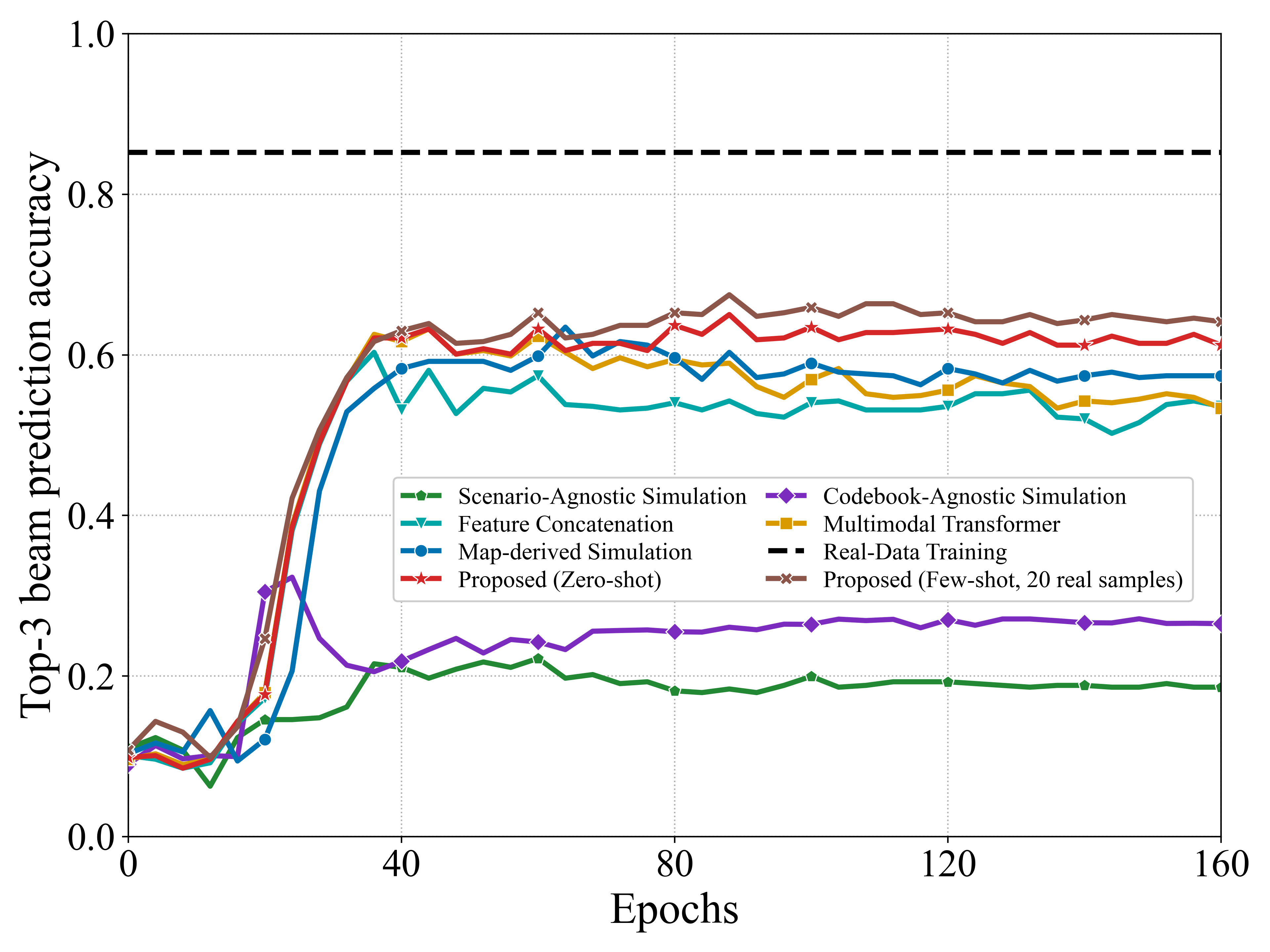}%
  \label{fig:beam_s3_top3}}
\hfill
\subfloat[Top-5 accuracy.]{%
  \includegraphics[width=0.49\textwidth]{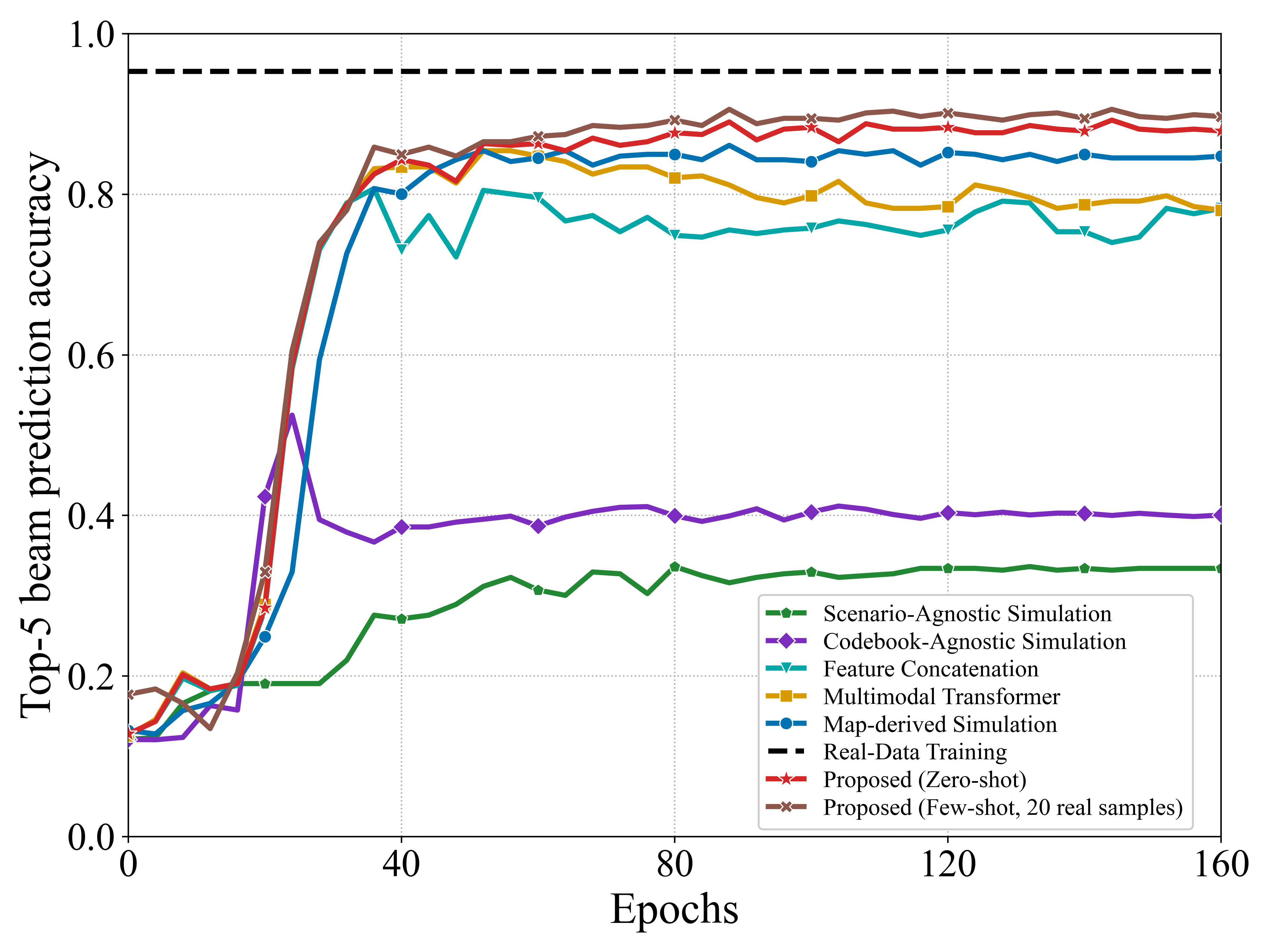}%
  \label{fig:beam_s3_top5}}
\caption{{\color{black}Beam prediction accuracy versus training epochs in Scenario 3: (a) Top-3 and (b) Top-5. The proposed few-shot model uses 20 real training samples. The real-data training result is a fixed reference.}}
\label{fig:beam_s3_metrics}
\label{fig:beam_s3_epoch}
\label{fig:exp_learning}
\end{figure*}

\begin{itemize}
    \item \textbf{Scenario-agnostic simulation:}
    Synthetic data are generated in a generic scene without target-specific geographic reconstruction.
    {\color{black}The sensing configuration and transmit codebook are matched
    to AIMS. The MoE model and synthetic data volume are also matched.}

    \item \textbf{Map-derived simulation:}
The target site is reconstructed from OpenStreetMap with buildings
represented by plain volumetric models. The communication vehicle follows
predefined straight-line trajectories with background traffic disabled.
Blender renders the RGB observations, and Sionna RT generates wireless
channels from the same prescribed scene states. The implementation uses
data-generation components adapted from~\cite{li2026multimodalnf}.

    \item \textbf{Codebook-agnostic simulation:}
    {\color{black}An oversampled discrete Fourier transform (DFT) transmit codebook replaces the target-device codebook when generating synthetic beam supervision.}
    {\color{black}Beam indices are aligned with the real codebook through a fixed mapping defined by codebook metadata.}
    This baseline retains the reconstructed scene and MoE model and applies only to beam prediction.

    \item \textbf{Feature concatenation:}
    Modality-specific features are concatenated and projected to the task head, replacing the MoE fusion mechanism.

    \item \textbf{Multi-modal transformer:}
    Transformer attention captures cross-modal interactions and generates fused features for task prediction.

    \item \textbf{Real-data training benchmark:}
    {\color{black}The same task-specific MoE model is trained directly on the complete real training partition.}
    This setting provides an empirical performance reference with a larger real-data budget.
\end{itemize}

\subsubsection{Agent Settings}

The agentic orchestration of AIMS is implemented with a locally deployed
Qwen3-14B~\cite{yang2025qwen3technicalreport} language model as the common LLM backbone. The LLM interprets
target-deployment requests, reasons over the structured domain knowledge and
current experiment state, and selects supported capabilities according to the
task requirements and intermediate-product dependencies. Physical solvers execute the
simulations, and numerical learners optimize the task models.

We construct 2 benchmark tiers for evaluating the shared orchestration
mechanism. Each tier contains 30 expert-designed canonical cases expressed
as 70 natural-language requests. In each tier, 20 cases are represented
by 3 paraphrases, while 10 additional cases use 1 formulation each.
The normal tier contains explicit and feasible deployment requests, with
8 task-endpoint cases, 16 dependency-and-reuse cases, and 6
recovery cases. {\color{black}The challenging tier introduces indirect endpoint
descriptions, implicit outputs, and artifact-provenance conflicts. It also
includes text-only constraints, unsupported goals, and unseen execution
failures. The tier contains 5 task-endpoint cases, 9 dependency-and-reuse
cases, and 5 constraint cases. It also contains 5 unsupported-request
cases and 6 recovery cases.} Across both
tiers, the benchmark contains 60 canonical cases and 140 natural-language
requests, resulting in 420 method-request evaluations across the 3
compared methods.

We compare the proposed framework with the following
baselines.
\begin{itemize}
    \item \textbf{Direct LLM:}
    {\color{black}The LLM receives the common capability catalog and a serialized
    experiment-state snapshot. It produces a single response without tool calls,
    validation evidence, or structured domain knowledge.}
    \item \textbf{Tool-only AIMS:}
    {\color{black}The method uses the same state interface and validator as
AIMS. It also uses the same feedback format and retry budget. The structured
domain knowledge is omitted.} It can call the corresponding
functions, which return structured schema, dependency, and
capability-constraint feedback.
\end{itemize}

\subsection{Performance Comparison for Beam Prediction}
\label{subsec:exp_tasks}
In this subsection, we evaluate beam prediction in Scenarios 3 and 4. The proposed zero-shot model uses no target-domain samples for adaptation, while the few-shot model uses 20 real training samples. The real-data training reference uses the complete real training partition.

\begin{figure*}[t]
\centering
\subfloat[Top-3 accuracy.]{%
  \includegraphics[width=0.49\textwidth]{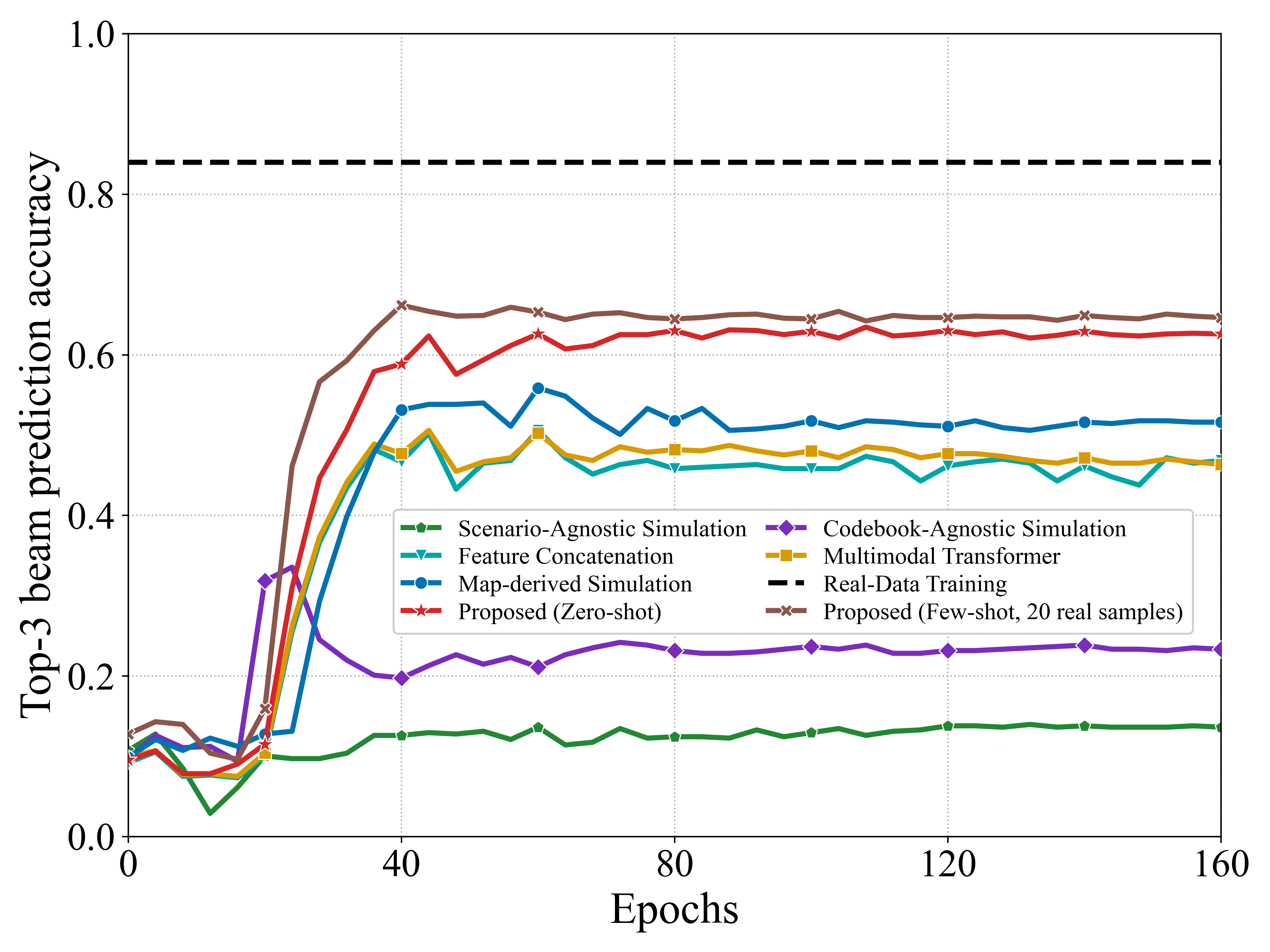}%
  \label{fig:beam_s4_top3}}
\hfill
\subfloat[Top-5 accuracy.]{%
  \includegraphics[width=0.49\textwidth]{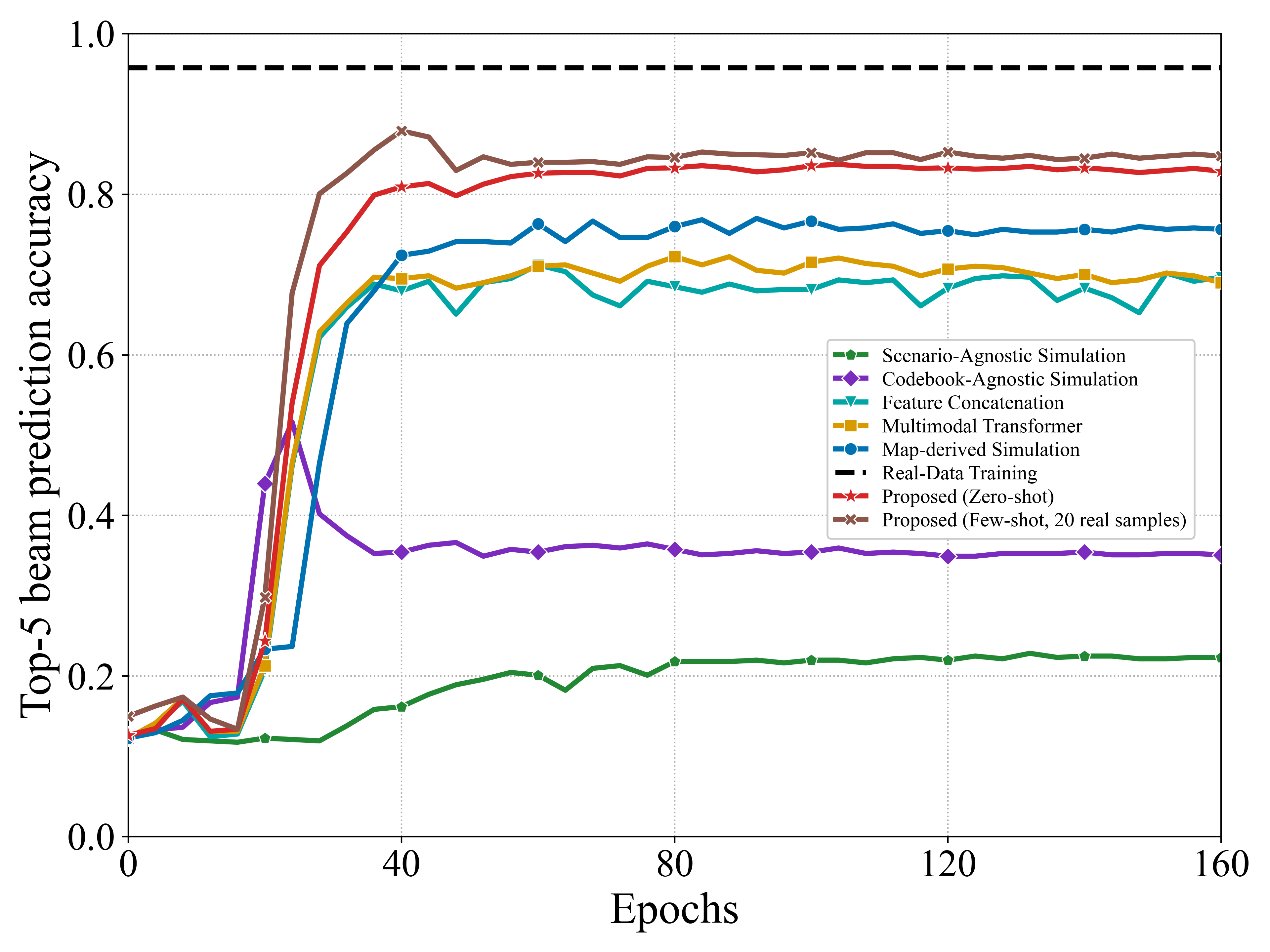}%
  \label{fig:beam_s4_top5}}
\caption{{\color{black}Beam prediction accuracy versus training epochs in Scenario 4: (a) Top-3 and (b) Top-5. The proposed few-shot model uses 20 real training samples. The real-data training result is a fixed reference.}}
\label{fig:beam_s4_metrics}
\label{fig:beam_s4_epoch}
\end{figure*}

Top-$k$ accuracy measures how often the best beam is included among the $k$ highest-ranked predictions:
\begin{equation}
 A_k=\frac{1}{N_{\rm te}}\sum_{i=1}^{N_{\rm te}}
 \ind\!\left\{y_i^{\rm b}\in\widehat{\mathcal B}_{i,k}\right\},
 \qquad k\in\{1,3,5\},
 \label{eq:exp_topk}
\end{equation}
where $N_{\rm te}$ is the real test-set size, $y_i^{\rm b}$ is the best measured beam, and $\widehat{\mathcal B}_{i,k}$ contains the Top-$k$ predictions. Figs.~\ref{fig:beam_s3_metrics} and~\ref{fig:beam_s4_metrics} report Top-3 and Top-5 accuracy.

Both proposed configurations outperform all 5 baselines in both scenarios. The largest gaps occur against scenario-agnostic and codebook-agnostic simulation, supporting alignment of scene-dependent propagation and beam-label definitions with the target deployment. The map-derived simulation baseline also remains below AIMS in both scenarios, indicating that target-site reconstruction with simplified geometry and prescribed mobility does not fully reproduce the benefit of the deployment-conditioned construction. The zero-shot model also exceeds feature concatenation and {\color{black}multi-modal transformer}. Its final Top-3 gain over the stronger fusion baseline is approximately 7 percentage points in Scenario 3 and 16 percentage points in Scenario 4.

Using 20 real training samples further improves all 4 plotted Top-$k$ results. {\color{black}The full real-data training reference remains higher. The gap is smaller for Top-5 than for Top-3, indicating stronger coverage when a larger candidate list is allowed.}

\subsection{Performance Comparison for Object Detection}
\label{subsec:exp_sensing}
In this subsection, we evaluate sensing through bounding-box detection of the communication vehicle from RGB and GPS observations. Because multiple vehicles may appear in the camera view, GPS provides positional context for associating the communication vehicle with the corresponding visual target. Scenarios 3 and 9 use zero-shot and 20-sample few-shot settings.

We evaluate detection using average precision (AP) over multiple
intersection-over-union (IoU) thresholds and AP at an IoU threshold of 0.50
(AP50). Let $\operatorname{AP}_u$ denote average precision from the
precision--recall curve at threshold $u$. Then
\begin{equation}
 \operatorname{AP50}=\operatorname{AP}_{0.50},\qquad
 \operatorname{AP}=\frac{1}{10}\sum_{j=0}^{9}
 \operatorname{AP}_{0.50+0.05j}.
 \label{eq:exp_detection_metrics}
\end{equation}
{\color{black}AP50 uses an IoU threshold of 0.50. AP averages thresholds from 0.50 to 0.95 to assess performance under stricter localization requirements.}

\begin{figure*}[t]
\centering
\subfloat[AP.]{%
  \includegraphics[width=0.49\textwidth]{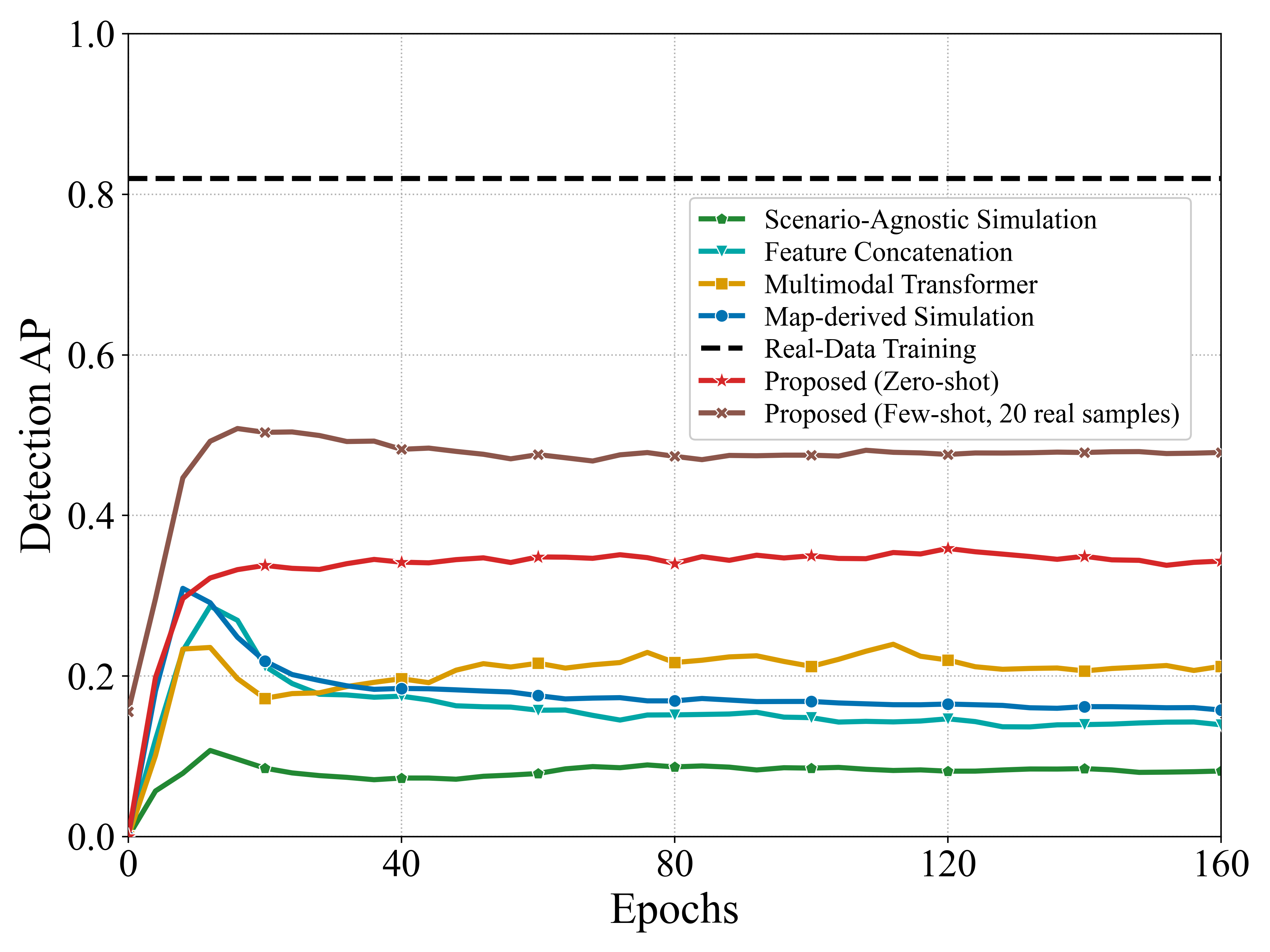}%
  \label{fig:det_s3_ap}}
\hfill
\subfloat[AP50.]{%
  \includegraphics[width=0.49\textwidth]{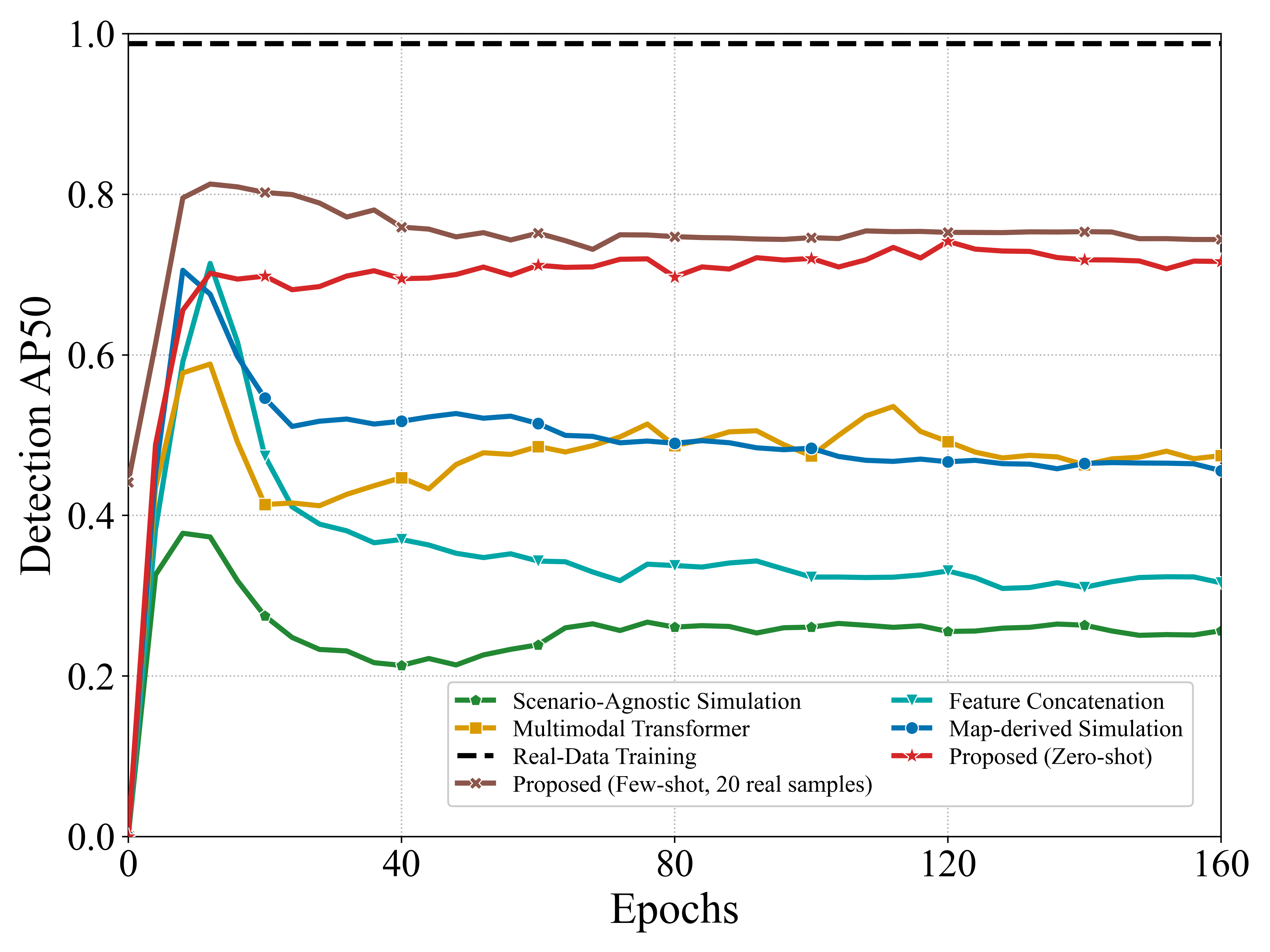}%
  \label{fig:det_s3_ap50}}
\caption{{\color{black}Detection performance in Scenario 3: (a) AP and (b) AP50. The proposed few-shot model uses 20 real training samples. The real-data training result is a fixed reference.}}
\label{fig:det_s3_metrics}
\label{fig:det_s3_epoch}
\end{figure*}

\begin{figure*}[t]
\centering
\subfloat[AP.]{%
  \includegraphics[width=0.49\textwidth]{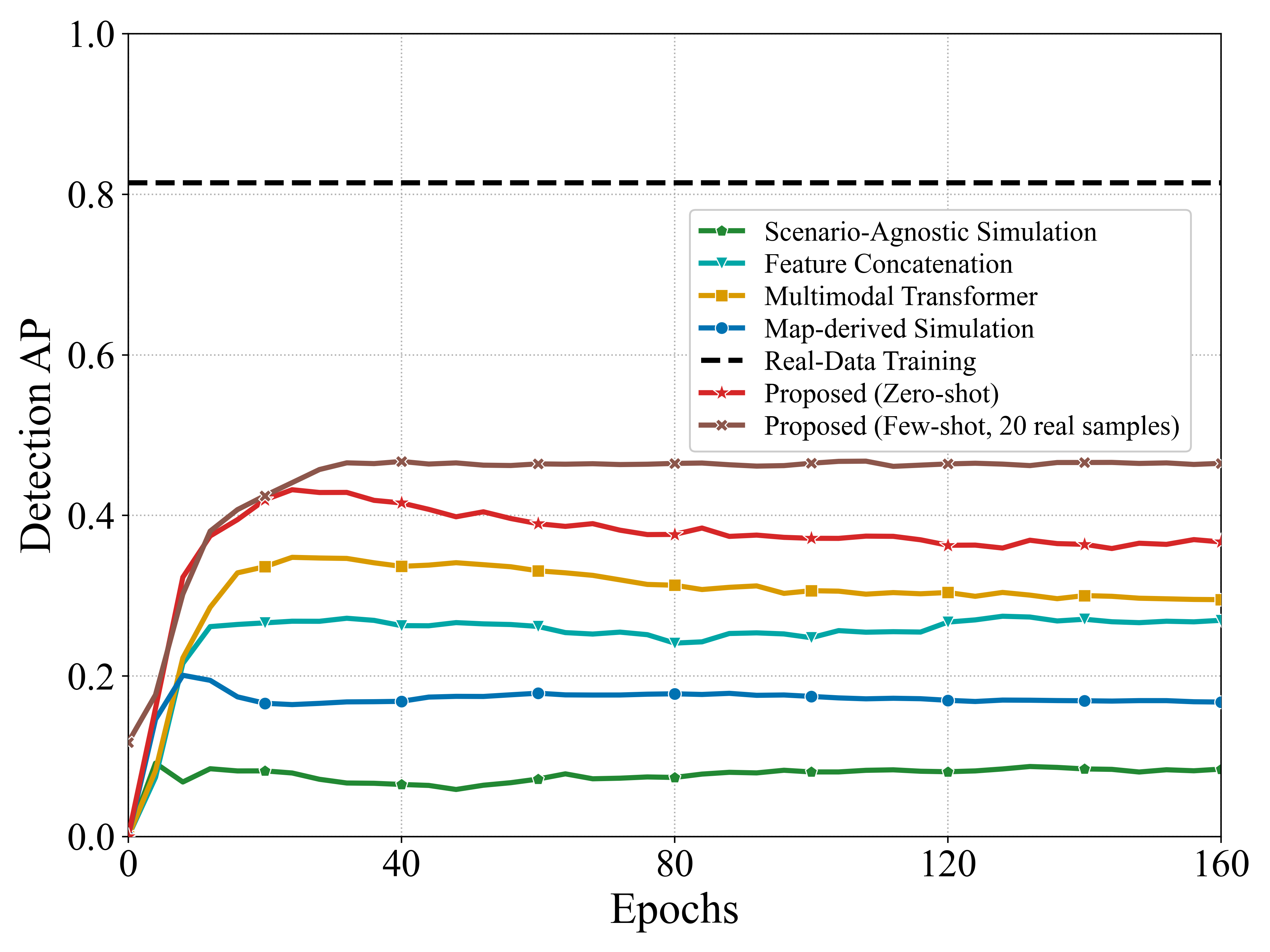}%
  \label{fig:det_s9_ap}}
\hfill
\subfloat[AP50.]{%
  \includegraphics[width=0.49\textwidth]{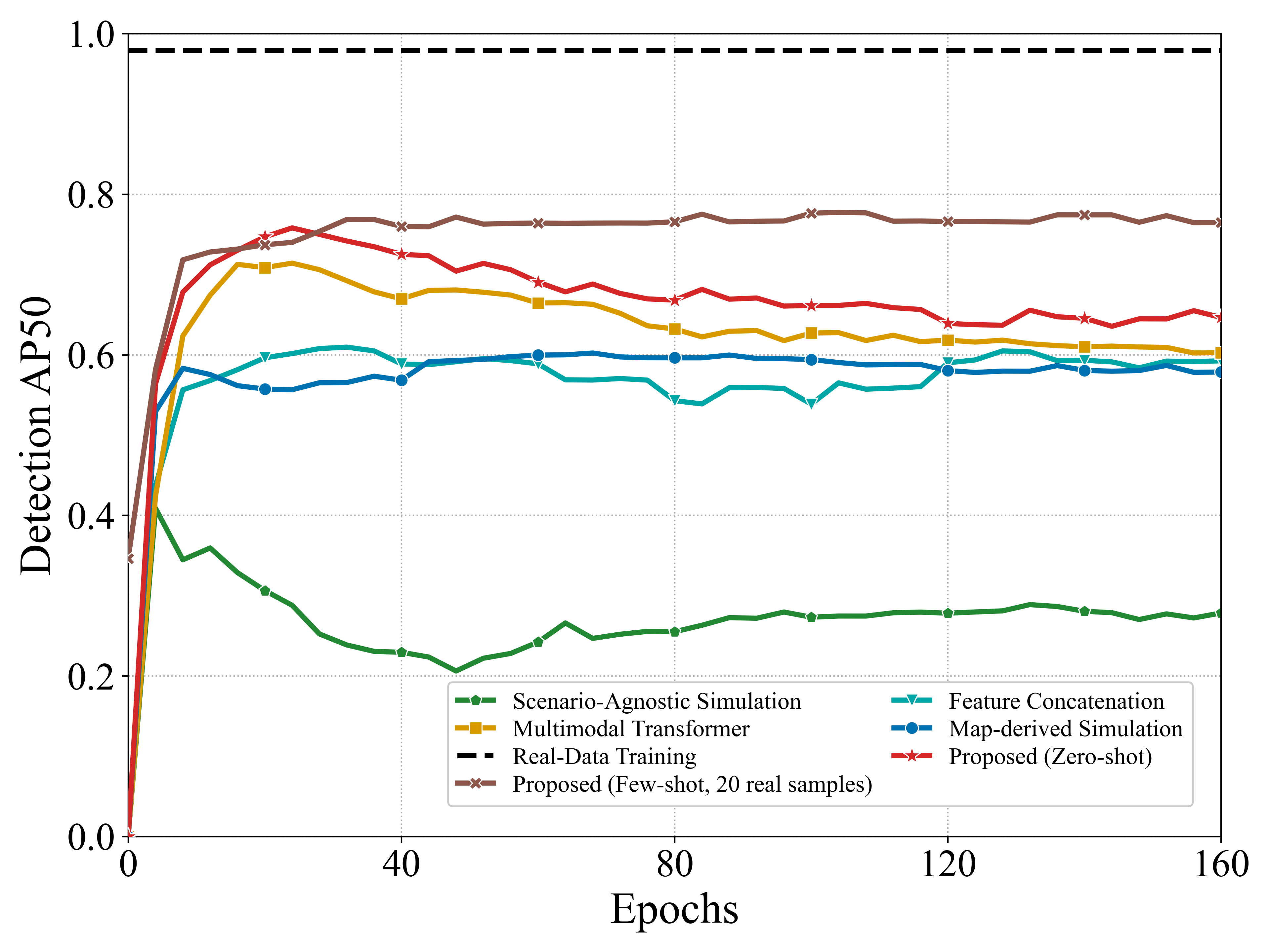}%
  \label{fig:det_s9_ap50}}
\caption{{\color{black}Detection performance in Scenario 9: (a) AP and (b) AP50. The proposed few-shot model uses 20 real training samples. The real-data training result is a fixed reference.}}
\label{fig:det_s9_metrics}
\label{fig:det_s9_epoch}
\end{figure*}

At the final epoch, the proposed zero-shot model exceeds scenario-agnostic simulation and both fusion baselines on AP and AP50 in both scenarios. Its AP advantage over the stronger fusion baseline ranges from approximately
6 to 13 percentage points across the two scenarios. The scenario-agnostic gap supports target-relevant scene construction, and the fusion comparisons support the proposed {\color{black}multi-modal learning configuration}. The map-derived simulation baseline also remains below the proposed models on both AP and AP50, providing a reference for the contribution of the deployment-conditioned scene construction beyond target-site map reconstruction alone. The codebook-agnostic simulation baseline is excluded because it changes wireless supervision without changing detection inputs or labels.

Few-shot learning with 20 real samples further improves both AP and AP50, and the advantage is maintained through the later training epochs. The gains therefore extend beyond detection at moderate box overlap to the more demanding localization criteria included in AP. A gap to the full real-data training reference remains, particularly in AP, indicating residual sim-to-real mismatch.

\subsection{Performance Comparison for Agentic Orchestration}

In this subsection, we evaluate task interpretation and feedback-driven
plan revision across diverse deployment requests. The expert-defined
contracts assess whether the selected capabilities and their dependencies
satisfy the task and execution conditions
in~\eqref{eq:aims_joint_problem}.

Following recent agentic wireless evaluation that considers task success,
first-try success, and iterative solution attempts
~\cite{li2026comagent}, we evaluate AIMS using task-level success and
fine-grained orchestration metrics. Plan task success rate (Plan-TSR) is the
fraction of requests whose final plan satisfies the expert-defined contract.
{\color{black}The contract covers task interpretation, target endpoint, and
capability selection. It also covers intermediate-artifact handling and
required recovery behavior.} Constraint satisfaction rate (CSR) is the fraction
of final plans that satisfy schema and capability constraints together with
dependency and cross-field constraints. {\color{black}It measures structural plan validity.} Artifact-dependency F1 is the
macro-averaged F1 score for determining whether each intermediate artifact
should be preserved, recomputed, or excluded from the requested workflow.
Capability-F1 measures agreement between the selected capabilities and those
required by the expert annotation. First-try success is the fraction of
initial plans satisfying the expert-defined contract. Avg. Attempts gives the
mean number of planning and replanning attempts per request.

\begin{table*}[t]
\centering
\caption{Agentic orchestration performance on 70 normal and
70 challenging natural-language deployment requests.}
\label{tab:agentic_orchestration}
\setlength{\tabcolsep}{4.5pt}
\renewcommand{\arraystretch}{1}
\resizebox{\textwidth}{!}{%
\begin{tabular}{llcccccc}
\toprule
Tier & Method
& Plan-TSR (\%)
& CSR (\%)
& Dependency-F1 (\%)
& Capability-F1 (\%)
& First-Try (\%)
& Avg. Attempts \\
\midrule

\multirow{3}{*}{Normal}
& Direct LLM
& 60.0
& \textbf{100.0}
& 88.1
& 82.2
& 60.0
& 1.000 \\

& Tool-only AIMS
& 72.9
& 98.6
& 87.0
& 77.4
& 52.9
& 1.286 \\

& \textbf{AIMS}
& \textbf{100.0}
& \textbf{100.0}
& \textbf{100.0}
& \textbf{100.0}
& \textbf{78.6}
& 1.229 \\

\specialrule{1.0pt}{2pt}{2pt}

\multirow{3}{*}{Challenging}
& Direct LLM
& 48.6
& 97.1
& 63.6
& 61.3
& 48.6
& 1.000 \\

& Tool-only AIMS
& 60.0
& 98.6
& 72.5
& 65.3
& 45.7
& 1.471 \\

& \textbf{AIMS}
& \textbf{88.6}
& \textbf{100.0}
& \textbf{92.4}
& \textbf{90.3}
& \textbf{68.6}
& 1.257 \\

\bottomrule
\end{tabular}%
}
\end{table*}

As shown in Table~\ref{tab:agentic_orchestration}, AIMS achieves
100\% Plan-TSR on the normal tier, exceeding Tool-only AIMS and Direct LLM
by 27.1 and 40.0 percentage points, respectively. The controlled comparison with Tool-only AIMS provides direct evidence for
the contribution of structured domain knowledge. {\color{black}The two methods
share the same model, capability interfaces, and state representation. They
also share the validator, feedback format, and retry budget. Their difference
is the availability of structured domain knowledge.} Incorporating this
knowledge increases Dependency-F1 from 87.0\% to 100\% and Capability-F1 from
77.4\% to 100\%. AIMS also achieves 78.6\% first-try success, compared with
52.9\% for Tool-only AIMS. {\color{black}Its average number of attempts is
1.229, compared with 1.286 for Tool-only AIMS.}

Under challenging requests, AIMS attains 88.6\% Plan-TSR, compared with
60.0\% for Tool-only AIMS and 48.6\% for Direct LLM. Its Dependency-F1 and
Capability-F1 remain 92.4\% and 90.3\%, respectively. {\color{black}Tool-only
AIMS achieves 72.5\% and 65.3\%.} The first-try success of AIMS decreases from
78.6\% on the normal tier to 68.6\%, and its average number of attempts rises
slightly from 1.229 to 1.257.

{\color{black}Compared with Tool-only AIMS, structured domain knowledge
improves first-try success and final-plan accuracy under shared feedback
settings. AIMS also achieves higher Plan-TSR with fewer attempts in both
tiers. Dependency-F1 and Capability-F1 measure agreement with execution
requirements. CSR measures structural validity, while Plan-TSR also includes
task interpretation and requested outcomes.}

\section{Conclusion}
\label{sec:conclusion}

In this paper, we presented AIMS for deployment-specific sim-to-real
multi-modal ISAC. AIMS grounds a natural-language deployment request in the
shared experiment state and uses structured domain knowledge to instantiate
the deployment-specific sim-to-real configuration, identify reusable
intermediate products, and generate the execution plan. The scene construction
agent realizes physically aligned sensing and wireless data under shared
physical states, while the scene understanding agent configures task-relevant
modalities, learning, and sim-to-real adaptation. Validation evidence enables
dependency-aware revision as deployment conditions change. Experiments on
DeepSense~6G demonstrate zero-shot inference and limited real-domain adaptation
for vehicle detection and beam prediction, while the orchestration benchmark
validates task interpretation, dependency handling, and feedback-driven
replanning across deployment requests. These results can further motivate
performance-driven closed-loop refinement, where downstream sensing and
communication performance can provide additional feedback for subsequent
configuration and adaptation decisions.

\bibliographystyle{IEEEtranAutoAbbr} 
\bibliography{references}

\end{document}